\documentclass[]{article}
\pdfoutput=1 
\usepackage{times}
\usepackage{fullpage}
\usepackage{courier}
\usepackage{url}
\usepackage{verbatim} 
\usepackage{graphicx,dsfont,subfig}
\usepackage{multirow}
\usepackage{algorithm}
\usepackage[noend]{algorithmic}
\usepackage{xspace}
\usepackage{amsmath}
\usepackage{amssymb}
\usepackage{amsthm}
\usepackage{empheq}
\usepackage{natbib}
\usepackage{lipsum}
\usepackage{color}
\usepackage{xspace}
\usepackage{paralist}
\usepackage{enumitem}
\usepackage{authblk}
\usepackage{url}

\def\prodd{\displaystyle\prod}

\def\prodd{\displaystyle\prod}

\def\Zb{\mathbf{Z}}

\def\zn{\mathbf{z}_{n}}

\def\Ib{\mathbf{I}}
\def\Xb{\mathbf{X}}
\def\xd{\mathbf{x}^d}
\def\x{\mathbf{x}}

\def\s2X{\sigma_x^2}
\def\Yb{\mathbf{Y}}
\def\yd{\mathbf{y}_{r}^d}
\def\ydnon{\mathbf{y}_{\neg nr}^d}
\def\Bb{\mathbf{B}}
\def\bd{\mathbf{b}^{d}}
\def\bdr{\mathbf{b}^{d}_r}
\def\s2B{\sigma_B^2}

\def\Ncal{\mathcal{N}}
\def\Ocal{\mathcal{O}}

\def\bb{\mathbf{b}}

\def\Pb{\mathbf{P}}
\def\Lb{\boldsymbol{\lambda}^d}

\newcommand{\xhdr}[1]{\vspace{1.7mm}\noindent{{\bf #1.}}}

\begin{document}

\title{General Latent Feature Models for Heterogeneous Datasets}

\author[1]{Isabel Valera\footnote{Corresponding author: miv24@cam.ac.uk}}
\author[2]{Melanie F. Pradier}
\author[1]{Maria Lomeli}
\author[1,3]{Zoubin Ghahramani}

\affil[1]{University of Cambridge, Cambridge, UK}
\affil[2]{Universidad Carlos III de Madrid, Madrid, Spain}
\affil[3]{Uber AI Labs, San Francisco, US}
\date{}
\begin{small}
\maketitle
\end{small}

\maketitle

\begin{abstract}
Latent feature modeling allows capturing the latent structure responsible for generating the observed properties of a set of objects. It is often used 
to make predictions either for new observations or for missing information in the original data, as well as to perform exploratory data analysis. 
However, although there is an extensive literature on latent feature models for homogeneous datasets,  where all the attributes that describe each object are of the same (continuous or discrete) nature,
there is a lack of work  on latent feature modeling for heterogeneous databases. 
 In this paper, we introduce a general Bayesian nonparametric latent feature model suitable for heterogeneous datasets, where the attributes describing each object can be either discrete, continuous or mixed variables. 
The proposed model presents several important properties. Firstly, it accounts for heterogeneous data while keeping the properties of conjugate models, which allows to infer the model in linear time with respect to the number of objects and attributes.
Secondly, its Bayesian nonparametric nature allows us to automatically infer the model complexity from the data, i.e., the number of features necessary to capture the latent structure in the data. 
Thirdly, the latent features in the model are binary-valued, this facilitates the interpretability of the obtained latent features in exploratory data analysis. Finally, a software package called GLFM toolbox is made publicly available for other researchers to use and extend. 
We show the flexibility of the proposed model by solving both prediction and data analysis tasks on several real-world datasets. 

\end{abstract}

\section{Introduction}
\label{sec:intro}
One of the aims of unsupervised learning is recovering the latent structure responsible for generating the observed properties or attributes of a set of objects. In particular, in latent feature modeling, one or more attributes of each object can be represented by an unobserved vector of latent features. It is assumed that the observations are generated from a distribution determined by those latent feature values. 
%
%
In other words, latent feature models allow us to represent, with only a few features, the immense redundant information present in the observed data, by 
capturing the statistical dependencies among the different objects and attributes. As a consequence,  they have been broadly used to make predictions either for new values of interest or missing information in the original data \cite[]{BPMF_2008, Fran2014}, as well as to perform exploratory data analysis, i.e, to better understand the data \cite[]{Blanco2012, valera2016infinite}. 
%

%
There is an extensive literature in latent feature modeling for homogeneous data, where all the attributes describing each object in the database are of the same type, that is continuous or discrete. In particular, these works assume that databases contain only either continuous data, usually modeled as Gaussian variables \cite[]{IBP, Todeschini2013}, or discrete, that can be either modeled by discrete likelihoods \cite[]{Li2009, PsiquiatrasJMLR, Fran2014} or simply treated as Gaussian variables \cite[]{BPMF_2008, Blanco2012, Todeschini2013}.  However, there still exists a lack of work dealing with heterogeneous databases, which in fact are common in real applications.  
For instance, the Electronic Health Records of hospitals might contain lab measurements (often real-valued or positive real-valued data), diagnoses (categorical data) and genomic information (ordinal, count data and categorical data). Another example are surveys, which often contain diverse information about the participants such as age (count data), gender (categorical data), salary (positive real data), etc. 
Despite the diversity of data types, the standard approach when dealing with heterogeneous datasets is to treat all the attributes, either continuous or discrete, as  Gaussian variables.

This paper presents a general latent feature model (GLFM) suitable for heterogeneous datasets, where the attributes describing each object can be either discrete, continuous or mixed variables. Specifically, we account for real-valued and positive real-valued as examples of continuous variables, and categorical, ordinal and count data as examples of discrete variables. 
The proposed model extends the essential building block of Bayesian nonparametric latent feature models, the Indian Buffet Process (IBP) by \cite{IBP}, to account for heterogeneous data while maintaining the model complexity of conjugate models.  
The IBP places a prior distribution over binary matrices where the number of columns, corresponding to latent features, is potentially infinite and can be inferred from the data. 
Among all the available latent feature models in the literature, we opt for the IBP due to two main reasons.  
First, the nonparametric nature of the IBP allows us to  automatically infer the appropriate model complexity,  i.e., the number of necessary features, from the data.
Second,  the IBP considers binary-valued latent features which has been shown to provide more interpretable results in data exploration than standard real-valued latent feature models~\cite[]{SuicidasNIPS,PsiquiatrasJMLR}. 
The standard IBP assumes real-valued observations combined with conjugate likelihood models that allow for fast inference algorithms~\cite[]{AcceleratedGibbs}. However, we handle heterogeneous databases, such that conjugacy is not preserved.
%

In order to propose a general observation model for the IBP that accounts for heterogeneous data while keeping the properties of conjugate models, we exploit two key ideas. First, we introduce an auxiliary real-valued variable (also called \emph{pseudo-observation}), such that, conditioned on it, the model behaves as the standard linear-Gaussian IBP in \cite[]{IBP}.  Second,  we assume that there exists a  function that transforms the pseudo-observation into an actual observation, mapping the real line into the (discrete or continuous) observation space  of each attribute in the data. 
These two key ideas allow us to derive an efficient inference algorithm based on collapsed Gibbs sampling, which presents linear complexity with the number of objects and attributes in the data. 

We show the flexibility and applicability of the proposed model by solving both prediction and data exploration tasks in several real-world datasets. 
In particular,  we first use the proposed model to estimate and replace missing data in heterogeneous databases (where the data is missing completely at random), showing that our approach for missing data estimation outperforms, in terms of accuracy, the Bayesian probabilistic matrix factorization (BPMF)~\cite[]{BPMF_2008} and the standard IBP which assumes Gaussian observations. These results have been previously discussed in \cite[]{Valera_NIPS2015}, where the main focus of the paper was missing data estimation (a.k.a. table completion). 
In contrast, this extended version of the paper focuses on the model itself, providing the necessary details on the GLFM and its inference to perform latent feature modeling in heterogeneous datasets, which  is a powerful tool not only for missing data estimation but also for exploratory data analysis tasks.  
%
This motivates the second part of the experiments, where 
 we present several examples of how to use the proposed model for data exploration in real-world datasets gathered from diverse application domains related to medicine, psychiatry, sociology and politics. 


The rest of the paper is organized as follows. In Section 2, we provide the details on the general Bayesian nonparametric latent feature model for heterogeneous datasets. In Section 3, we develop an  inference algorithm based on the Gibbs sampler, where we make use of the introduced pseudo-observation to collapse the sampler. In Section 4, we apply our model to two real-world applications, missing data estimation and data analysis. Finally, Section 5 is devoted to the discussion on the proposed latent feature model for heterogeneous datasets and its potential applications.

\section{Latent Feature Model for Heterogeneous Data}
\label{sec:model}
We assume that the data can be stored in an observation matrix $\Xb$ of size $N\times D$, each of the $N$ objects is defined by a set of $D$ attributes. Let $x^d_{n}$ denote each entry of the observation matrix $\Xb$, which might be of the following types:
\begin{compactitem}
\item Continuous variables:
\begin{compactenum}
\item Real-valued, $x_n^d\in \Re$ 
\item Positive real-valued, $x_n^d\in \Re_+$. 
\end{compactenum}
\item Discrete variables:
\begin{compactenum}
\item Categorical data, $x_n^d$ takes a value in a finite unordered set, e.g., $x_n^d\in\{$`blue', `red',  `black'$\}$.
\item Ordinal data, $x_n^d$ takes values in a finite ordered set, e.g., $x_n^d\in\{$`never',  `sometimes', `often', `usually', `always'$\}$.
\item Count data, $x_n^d\in \{0, \ldots, \infty\}$. 
\end{compactenum}
\end{compactitem}
As in standard latent feature models, we assume that  $x^d_{n}$ can be explained by a $K$-length vector of latent features associated to the $n$-th data point, $\zn=[z_{n1},\ldots,z_{nK}]$, and a weight vector\footnote{For convenience, we capitalized here the notation for the weight vectors $\Bb^d$.} $\Bb^d=[b^d_{1},\ldots,b^d_{K}]$ ($K$ being the number of latent variables), whose elements $b^d_k$ weight the contribution of the $k$-th latent feature to the $d$-th dimension of $\Xb$. 
Under this assumption, the likelihood can be factorized as
\begin{equation*}
p(\Xb| \Zb, \{\Bb^d, \Psi^d\}_{d=1}^D)
= \prodd_{d=1}^D  \prodd_{n=1}^N p(x^d_n| \zn, \Bb^d, \Psi^d), 
\end{equation*}
where  $\Psi^d$ denotes the set of random variables necessary to define the distribution of the $d$-th attribute. 
%
%
Here, we assume binary-valued latent binary feature vectors $\zn$, which are gathered in an $N\times K$ matrix $\Zb$ that follows an IBP prior with concentration parameter $\alpha$, denoted by $\Zb \sim \mathrm{IBP}(\alpha)$ \cite[]{IBP}. Additionally, we place a Gaussian distribution with zero mean and covariance matrix $\s2B\Ib_K$ over the weight vectors $\Bb^d$. 
%
%
Note that if $x^d_{n} \in \Re$ is assumed to be Gaussian with mean $\zn \Bb^d$ for the $d=1,\ldots, D$ attributes, the above model is equivalent to the  standard IBP with Gaussian observations~\cite[]{IBP}, therefore can be efficiently learnt using the properties of the Gaussian distribution~\cite[]{AcceleratedGibbs}. 
However, under heterogeneous (or non-Gaussian) observation matrices, developing an efficient inference algorithm is not straightforward, since the advantages of conjugate priors do not hold in the general case.

\begin{figure}[t]
\includegraphics[width=0.5\textwidth]{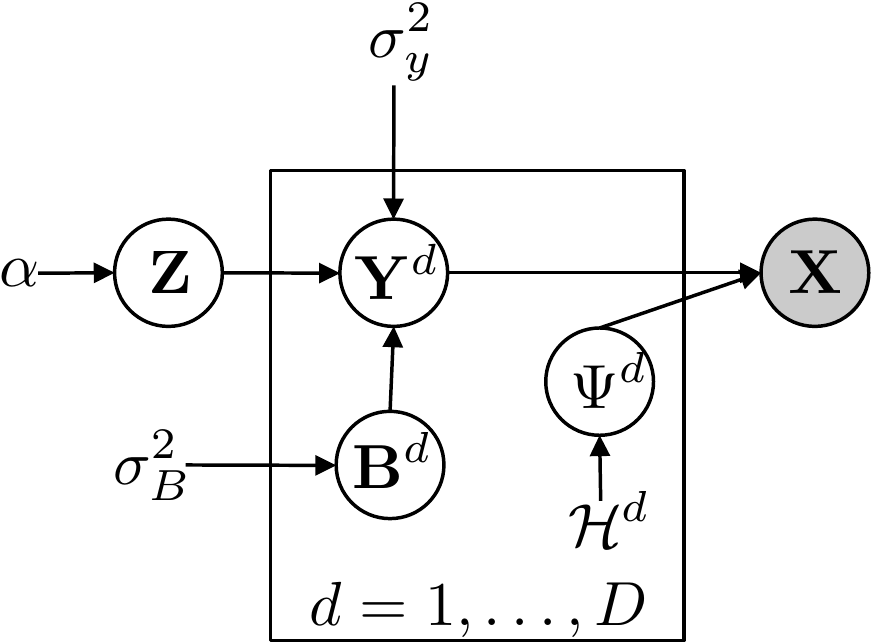}
\centering
\caption{\textbf{Graphical Model for the Generalized Latent Feature Model.} Grey nodes represent observed variables, white nodes correspond to latent variables. Introducing the pseudo-observations $\mathbf{Y}^d$ allow us to deal with heterogeneous data.}
\label{fig:IBPdiscreto}
\end{figure}

To solve this limitation, we introduce an auxiliary Gaussian variable $y_n^d$ (which we might refer to as \emph{pseudo-observation}) per entry  $x_n^d$ in the observation matrix, and assume that there exists a transformation function $f_d(\cdot)$ over the variables $y_{n}^d$ to obtain the observations $x_n^d$, mapping the real line $\Re$ into the observation space  of the $d$-th attribute in the observation matrix $\Omega_d$, i.e.,
\begin{equation}
\begin{array}{cccc}
f_d: &\Re  &\mapsto & \Omega_d\\
&y_n^d &\to  & x_n^d
\end{array}.
\vspace{-2mm}
\end{equation} 
Here, we assume that $y_{n}^d$ is Gaussian distributed with mean $\zn\Bb^d$ and variance $\sigma^2_y$, i.e.,
\begin{equation*}
p(y_{n}^d|\zn,\Bb^d)=\Ncal(y_{n}^d|\zn \Bb^d,\sigma^2_y),
\end{equation*}
such that, when conditioned on the pseudo-observations, the latent variable model behaves as a standard IBP with Gaussian observations. 
Additionally, Section~\ref{sec:likFunctions} details the mapping functions from the real line $\Re$ into each of the discrete and continuous spaces.%
Using auxiliary Gaussian variables to link a latent model with the observations has been previously used in Gaussian processes for multi-class classification \cite[]{Girolami05} and for ordinal regression \cite[]{Chu2005}. However, to our knowledge, this simple approach has not been used to account for mixed continuous and discrete data. The existent approaches for the IBP with discrete observations propose  non-conjugate likelihood models and approximate inference algorithms \cite[]{SuicidasNIPS, PsiquiatrasJMLR,valera2016infinite}. 

The resulting generative model is shown in Figure~\ref{fig:IBPdiscreto}, where $\Zb$ is the IBP latent matrix, and $\Yb^d$ and $\Bb^d$ contain, respectively, the pseudo-observations $y_{n}^d$ and the weight factors $b^d_k$ for the $d$-th dimension of the data. Additionally, $\Psi^d$ denotes the set of auxiliary random variables needed to obtain the observation vector $\xd$ given $\Yb^d$, and $\mathcal{H}^d$ contains the hyper-parameters associated to the random variables in $\Psi^d$.  
In order to obtain more interpretable results while performing data exploration, we also assume that the latent feature matrix $\Zb$ might be extended to contain an extra latent feature that is active for every object in the data, playing the role of a bias term similar to \cite[]{SuicidasNIPS, PsiquiatrasJMLR,valera2016infinite}.  



\subsection{Mapping Functions}\label{sec:likFunctions}
In this section, we define the set of functions that transforms the pseudo-observations $y_{n}^d$ into the corresponding observations $x_n^d$, i.e., that maps from the real line $\Re$ to the (continuous or discrete) observation space of the $d$-th attribute describing the data. Since each attribute (dimension) in $\Xb$ may contain any discrete or continuous data types, we provide a mapping function for each kind of data and the corresponding likelihood function for heterogeneous data.

\subsubsection{Continuous Variables}
In the case of continuous variables, we  assume that the mapping functions are of the form $x=f(y+u)$, where $f(\cdot)$ is a continuous invertible and differentiable function and $u$ corresponds to additive Gaussian noise with variance $\sigma^2_u$. In such case, we can obtain the corresponding likelihood function (after integrating out the pseudo-observation $y_{n}^d$) as 
\vspace{-2mm}
\begin{equation}
 p(x^d_n |\zn,\Bb^d)= \frac{1}{\sqrt{2\pi (\sigma^2_y +\sigma^2_u)}}\exp\left\{-\frac{1}{2 (\sigma^2_y +\sigma^2_u)}(f^{-1}(x^d_n)-\zn\Bb^d)^2\right\} \left|\frac{d}{dx^d_n}f^{-1}(x^d_n)\right|,
\end{equation}
where $f^{-1}(\cdot)$ is the inverse function of the transformation $f(\cdot)$, i.e., $f^{-1}(f(v))=v$. Next, we provide examples of mapping functions that account for real-valued and positive real-valued data. 

\xhdr{Real-valued Data}
In order to obtain real-valued observations, i.e., $x_n^d\in \Re$, we need a transformation over $y_n^d$ that maps from the real numbers to the real numbers, i.e., $f_d: \Re \rightarrow \Re$. 
%
The simplest case is to assume that $x=f_d(y+u)= y+u$. Therefore, each observation is distributed as $x_n^d \sim \Ncal(\zn\bb_\Re^d,\sigma^2_y + \sigma^2_u)$. 
Nevertheless, other mapping functions can be used, e.g., one might opt for the transformation 
\vspace{-1mm}
\begin{equation*}
x=f_d(y+u)= w (y+u)+\mu,
\vspace{-1mm}
\end{equation*}
where $w$ and $\mu$ are parameters which allow the user to scale or shift the attribute. A common choice would be taking $w = 1 / \mathbb{Var}[\xd]$ and $\mu = \mathbb{E}[\xd]$, which normalise the data.
The corresponding the auxiliary variables and hyper-parameters are $\Psi^d = \{u_d^n\}$ and $\mathcal{H}^d=\{\sigma^2_{u}, w, \mu\}$.

\xhdr{Positive Real-valued Data}
In order to obtain positive real-valued observations, i.e., $x_n^d\in \Re_+$, we can apply any transformation over $y_n^d$ that maps the real numbers to the positive real numbers, i.e.,  $f_d: \Re \rightarrow \Re_+$, as long as $f_d$  is an invertible and differentiable function. 
%
%
An example of this kind of functions is:
\begin{equation*}
f_d(y) =  \log(\exp(wy + \mu)+1),
\end{equation*}
 where $w$ and $\mu$  are hyper-parameters. 
Similarly to the case of real-valued attributes, here we also make use of the Gaussian variable $u_n^d$  to obtain $x_n^d$ from $y_n^d$, therefore, $\Psi^d = \{u_d^n\}$ and $\mathcal{H}^d=\{\sigma^2_{u}, w, \mu\}$.

\subsubsection{Discrete Variables}
In the case of discrete variables, there is not a general way to map the real line into any type of discrete variable, therefore, we need to derive a different transformation for each of the considered types of discrete variables, i.e., categorical, ordinal and  count data. 

\xhdr{Categorical Data}
Let $x^d_n$ be a categorical observation, namely, it can take values in the index set given by $\{1,\ldots, R_d\}$. Hence, assuming a multinomial probit model, we can then write:
\begin{equation}
x_{n}^d= f_{d}(y_{n}^d) = \stackrel{\text{argmax}}{r\in\{1,\ldots, R_d\}} y_{nr}^d,
\end{equation}
with $y_{nr}^d  \sim \Ncal(y_{nr}^d|\zn\bd_r,\sigma^2_y)$ where $\bd_r$ denotes the $K$-length weight vector, in which each $b_{kr}^d$ measures the influence of the $k$-th feature for the observation $x^d_{n}$ taking value $r$.  Under this likelihood model, we have as many pseudo-observations $y^d_{nr}$ and weight vectors $\bd_r$ per observation as number of categories in the $d$-th attribute, i.e., $r\in\{1,\ldots, R_d\}$.  
In this case, the pseudo-observations can be gathered  in  the $N \times R_d$ matrix $\Yb^d$ and the weight factors in a $K \times R_d$ matrix $\Bb^d$. 

%
Under this observation model, we can write $y_{nr}^d=\zn\bdr+u^d_{nr}$, 
where $u^d_{nr}$ is a Gaussian noise variable with variance $\sigma^2_y$, therefore, we can obtain, as described in~\cite[]{Girolami05}, the probability of each element $x_n^d$ taking value $r\in \{1, \ldots, R_d\}$ as
\begin{equation}\label{eq:LikCat}
p(x_{n}^d =r |\zn,	\Bb^d)  = \mathbb{E}_{p(u)}\Bigg[\prodd_{\substack{j=1\\ j\neq r}}^{R_d}\Phi \Big(u + \zn(\bd_{r}-\bd_{j})\Big)\Bigg],
\end{equation}
where the subscript $r$ in $\bd_{r}$ refers to the column in $\Bb^d$ ($r \in \{1,\ldots, R_d\}$), $\Phi(\cdot)$ denotes the cumulative distribution function of the standard normal distribution and $\mathbb{E}_{p(u)}[\cdot]$ denotes expectation with respect to the distribution $p(u)=\Ncal(0,\sigma^2_{y})$. 
Thus, the auxiliary variables and hyper-parameters are defined as $\Psi^d = \{u_d^n\}$ and $\mathcal{H}^d=\{\sigma^2_{u}\}$.

\xhdr{Ordinal Data}
Consider ordinal data, in which each element $x_n^d$ takes values in the ordered index set $\{1,\ldots, R_d\}$. Then, assuming an ordered probit model, we can write
\begin{equation}
x_{n}^d= f_{d}(y_{n}^d) = \left\{
  \begin{array}{l l}
    1 & \quad \text{if } y_{n}^d \leq\theta^d_1\\
    2 & \quad \text{if } \theta^d_1 <y_{n}^d \leq\theta^d_2\\
     & \vdots \\
    R_d & \quad \text{if } \theta^d_{R_d-1} <y_{n}^d\\
  \end{array} \right.
\end{equation}
where again $y_{n}^d$ is Gaussian distributed with mean $\zn\Bb^d$ and variance $\sigma^2_y$, and $\theta^d_r$ for $r\in\{1,\ldots, R_d-1\}$ are the thresholds that divide the real line into $R_d$ regions.
We assume that the thresholds $\theta^d_r$ are sequentially generated from the truncated Gaussian distribution $\theta_r^d\propto\Ncal(\theta_r^d| 0, \sigma^2_{\theta}) \mathbb{I}(\theta_r^d > \theta_{r-1}^d)$, 
where $\theta^d_0=-\infty$ and $\theta^d_{R_d}=+\infty$. In this case, the value of $x_{n}^d$ is determined by the region in which $y_n^d$ falls and, as opposed to the categorical case, now we have a unique weight vector $\Bb^d$ and a unique Gaussian variable $y_n^d$ for each observation $x_{n}^d$. 

Under the ordered probit model \cite[]{Chu2005}, the probability of each element $x_n^d$ taking value $r\in \{1,\ldots, R_d\}$ can be written as
\begin{equation}\label{eq:LikCat}
p(x_{n}^d =r |\zn,	\Bb^d)  = \Phi \Bigg(\frac{\theta^d_r-\zn\Bb^d}{\sigma_y}\Bigg)-\Phi \Bigg(\frac{\theta^d_{r-1}-\zn\Bb^d}{\sigma_y}\Bigg).
\end{equation}
Let us finally remark that, if the $d$-th dimension of the observation matrix contains ordinal data, the set of auxiliary variables reduces to the thresholds $\Psi^d = \{ \theta^d_1,\ldots, \theta^d_{R_d-1}\}$, and thus, $\mathcal{H}^d=\{\sigma^2_{\theta}\}$. 

\xhdr{Count Data}
In the case of count data, each observation $x^d_{n}$ takes non-negative integer values, , $x^d_{n} \in \{0, \ldots, \infty\}$. 
Then, we assume 
\begin{equation}
x^d_{n}=  f_{d}(y_{n}^d) =\lfloor f_{\Re_+}(y_{n}^d) \rfloor,
\end{equation}
where $\lfloor v \rfloor$ returns the floor of $v$, that is the largest integer that does not exceed $v$, and $f_{\Re_+}: \Re \rightarrow \Re_+$ is an invertible function that maps the real numbers to the positive real numbers. We can therefore write the likelihood function as
\begin{equation}
 p(x^d_n |\zn,\Bb^d)=\Phi \Bigg(\frac{f^{-1}(x^d_n+1)-\zn\Bb^d}{\sigma_y}\Bigg)-\Phi \Bigg(\frac{f^{-1}(x^d_n)-\zn\Bb^d}{\sigma_y}\Bigg),
\end{equation}
where $f_{\Re_+}^{-1}: \Re_+ \rightarrow \Re$ is the inverse function of the transformation $f_{\Re_+}(\cdot)$. 
In this case, there are no auxiliary random variables $\Psi^d $ or hyper-parameters $\mathcal{H}^d$ which is specified by empty sets.

\section{Inference}
\label{sec:inference}
In this section, we describe our algorithm for learning the latent variables given the observation matrix. 
%
%
In order to learn jointly the latent vectors $\zn$, the weight factors $\Bb^d$, and the auxiliary variables $\Psi^d$, we use a Markov Chain Monte Carlo (MCMC) inference scheme. Such methods have been broadly applied to infer the IBP matrix, e.g., in \cite[]{IBP, Williamson, Titsias}.
The proposed inference algorithm, summarized in Algorithm~\ref{alg:inference}, exploits the information in the available data to learn similarities among objects (captured in our model by the latent feature matrix $\Zb$), and {{identifies}} how these latent features show up in the attributes that describe the objects (captured in our model by $\Bb^d$). 

In Algorithm~\ref{alg:inference}, we first update the latent matrix $\Zb$. Note that conditioned on $\{\Yb^d\}_{d=1}^{D}$, both the latent matrix $\Zb$ and the weight matrices $\{\Bb^d\}_{d=1}^D$ are independent of the observation mthereforeatrix $\Xb$. Additionally, since $\{\Bb^d\}_{d=1}^D$ and $\{\Yb^d\}_{d=1}^D$ are Gaussian distributed, we can marginalize out the weight matrices $\{\Bb^d\}_{d=1}^D$ to obtain $p(\{\Yb^d\}_{d=1}^D|\Zb)$. In order to learn the matrix $\Zb$, we apply the collapsed Gibbs sampler which presents better mixing properties than the uncollapsed version. As a consequence, is the standard method of choice in the context of the standard linear-Gaussian IBP~\cite[]{IBP}. However, this algorithm suffers from a high computational cost, its is cubic in the number of data points $N$ at every iteration, which is a prohibitive cost when dealing with large databases. We use of the accelerated Gibbs sampler~\cite[]{AcceleratedGibbs} as an alternative, fast albeit approximate, scheme for inference. This algorithm presents linear complexity with the number of objects $N$ in the observation matrix.

Second, we sample the  weight factors in $\Bb^d$, which is a $K \times R_d$ matrix in the case of categorical attributes, otherwise, it is a $K$-length column vector. We denote each column vector in $\Bb^d$ by $\bdr$. The posterior over the weight vectors is given by
\begin{equation}\label{eq:pB}
p(\bdr | \yd,\Zb) = \Ncal(\bdr|\Pb^{-1}\Lb_{r}, \Pb^{-1}),
\end{equation}
where $\Pb=\Zb^{\top}\Zb +1/\s2B\Ib_k$ and $\Lb_{r}=\Zb^{\top}\yd$, with $\yd$ the $r$-th column of $\Yb^d$. Here, $r$ takes values in $\{1, \ldots, R_d\}$ in the case of categorical observations, while $r=1$ for the rest of types of variables.  Note that the covariance matrix $\Pb^{-1}$ depend neither on the dimension $d$ nor on $r$, so we only need to invert the $K \times K$ matrix $\Pb$ once at each iteration. In Section~\ref{subsec:sampleZ}, we describe how to efficiently sample $\Zb$, as well as how to efficiently compute $\Pb$ after  the corresponding changes are made in the $\Zb$ matrix by rank one updates. Because of this, we manage to bypass the computation of the matrix product $\Zb^{\top}\Zb$.
%
Once we have updated $\Zb$ and $\Bb^d$, we sample each element in $\Yb^d$ from the distribution $\Ncal(y_{nr}^d| \zn \bd_r, \sigma^2_y)$ if the observation $x_n^d$ is missing, and from the posterior  $p(y_{nr}^d| x^d_n,\zn, \Bb^d)$ specified in Section~\ref{subsec:sampleY}, otherwise.
Finally, we sample the auxiliary variables in $\Psi^d$ from their posterior distributions (also detailed in the Section~\ref{subsec:sampleY}) if necessary\footnote{The set of auxiliary variables  for the $d$-dimension, $\Psi^d$, can be augmented to contain the variance of the pseudo-observations $\Yb^d$ associated to the $d$-th attribute, which we denote  by $\sigma_{d}^2$ and for which we assume an inverse-gamma prior with parameters $\beta_1$ and $\beta_2$. Under this prior distribution, the posterior of $\sigma_{d}^2$ is  an inverse-gamma with parameters $\beta_1+ NS_d/2$ and $\beta_2+ \sum_{n=1}^N \sum_{r=1}^{S_d} (y_{nr}^d- \zn \bb_r^d)/2$, where $S_d$ is equal to the number of categories $R_d$ for those dimensions $d$ that contain categorical attributes, and it is equal to $S_d=1$, otherwise.} This two latter steps involve, in the worst case, sampling from a doubly truncated univariate normal distribution, for which we make use of the algorithm in \cite[]{robert1995}. 
%

\begin{algorithm}[h]
\caption{Inference Algorithm.}
\label{alg:inference}
\begin{algorithmic}[1]
 \REQUIRE $\Xb$ 
 \ENSURE $\Zb$ and $\{\Yb^d\}_{d=1}^D$\\
 \FOR{ each iteration}
 	\STATE Update  $\Zb$  given $\{\Yb^d\}_{d=1}^D$ as detailed in Section~\ref{subsec:sampleZ}.
 	\FOR{$d=1,\ldots,D$}
 		\STATE Sample  $\Bb^d$  given  $\Zb$ and $\Yb^d$ according to (\ref{eq:pB}).
 		\STATE Sample  $\Yb^d$  given  $\Xb$, $\Zb$ and $\Bb^d$ as shown in Section~\ref{subsec:sampleY}.
 		\STATE Sample  $\Psi^d$ (if needed) as shown in Section~\ref{subsec:sampleY}.
 	\ENDFOR
 \ENDFOR 
 \hspace*{-19pt}
 \textbf{ Output}:  $\Zb$, $\{\Bb^d\}_{d=1}^D$ and $\{\Psi^d\}_{d=1}^D$ 
\end{algorithmic}
\end{algorithm}

\subsection{Details on the Accelerated Gibbs Sampler}\label{subsec:sampleZ}
In this section we we review and adapt the sampler in  \cite[]{AcceleratedGibbs}, where the authors presented a linear-time accelerated Gibbs sampler for conjugate IBP models that effectively marginalize out the weight factors. The per-iteration complexity of this algorithm is $\Ocal(N(K^2 + KD))$, which is comparable to the uncollapsed linear-Gaussian IBP sampler that has per-iteration complexity $\Ocal(NDK^2)$ but does not marginalize out the weight factors, and as a result, presents slower convergence rate. Next, we explain how to adapt this algorithm for the proposed IBP model for heterogeneous data. 

The accelerated Gibbs sampling algorithm exploits the Bayes rule to avoid the cubic complexity with $N$ due to the computation of the marginal likelihood in the collapsed Gibbs sampler. In particular, it applies the Bayes rule to obtain the probability of each element in the latent feature matrix $\Zb$ of being active  as
\begin{equation}\label{eq:pz_y}
\begin{split}
 p(z_{nk}=1|\{\Yb^d\}_{d=1}^{D},\Zb_{\neg nk} ) 
&\propto \frac{m_{\neg n,k}}{N} \prod_{d=1}^D \prod_{r=1}^{S_d}\int_{\bdr} p(y_{nr}^d | \zn,\bdr) p(\bdr|  \ydnon \Zb_{\neg n}) d\bdr,
\end{split}
\end{equation}
where $S_d$ is the number of columns in matrices $\Yb^d$ and $\Bb^d$ ($S_d$ is the number of categories $R_d$ for those dimensions $d$ that contain categorical attributes, and $S_d=1$ otherwise), $\Zb_{\neg n}$ corresponds to matrix $\Zb$ after removing the $n$-th row, vector $\ydnon$ is the $r$-th column of matrix $\Yb^d$ without the element $y_{nr}^d$, and $p(\bd_r | \xd_{\neg n},  \Zb_{\neg n})$
is the posterior of $\bd_r$ computed without taking the $n$-th datapoint into account, i.e.,
\begin{equation}
p(\bdr | \ydnon, \Zb_{\neg n})= \Ncal(\bdr | \Pb_{\neg n}^{-1}\Lb_{\neg n r}, \Pb_{\neg n}^{-1}),
\end{equation}
where $\Pb_{\neg n}=\Zb_{\neg n}^{\top}\Zb_{ \neg n}+1/\s2B\Ib_K$ and $\Lb_{\neg n y}=\Zb_{\neg n}^{\top}\ydnon$ are the natural parameters of the Gaussian distribution. In this case, we condition on the Gaussian pseudo-observations $\{\Yb^d\}_{d=1}^{D}$, instead of the actual observations $\Xb$, to compute the marginal distribution $ p(z_{nk}=1|\{\Yb^d\}_{d=1}^{D},\Zb_{\neg nk} )$.

In contrast to \cite[]{AcceleratedGibbs}, we use the natural parameterisation for the Gaussian distribution over the posterior of $\bdr$ instead of the mean and covariance matrix. This formulation allows us to compute the full posterior over the weight factors as 
\begin{equation}
p(\bdr | \yd, \Zb)= \Ncal(\bdr|\Pb^{-1}\Lb_r, \Pb^{-1}).
\end{equation}
$\Pb=\Pb_{\neg n}+\zn^{\top} \zn$ and $\Lb_r=\Lb_{\neg nr}+\zn^{\top}y_{nr}^d$ are the natural parameters of the Gaussian distribution. 

In the accelerated Gibbs sampling scheme, we iteratively sample the value of each element $z_{nk}$, after marginalizing out the weight factors $\Bb^d$, according to
\begin{equation}\label{eq:pz_yFin}
\begin{split}
 p(z_{nk}=1|\{\Yb^d\}_{d=1}^{D},\Zb_{\neg nk} ) 
&\propto \frac{m_{\neg n,k}}{N} \prod_{d=1}^D \prod_{r=1}^{S_d} \Ncal(y_{nr}^d| \zn \Lb_{\neg nr}, \zn \Pb_{\neg n}  \zn^{\top}+\sigma^2_y).
\end{split}
\end{equation}
 For each object $n$ we first sample the existing latent features $z_{nk}$ for $k= 1, \ldots, K_+$ (where $K_+$  is the number of non-zero columns in $\Zb$, or number of active features up to this iteration). Succesively, we sample the number of new features necessary to explain that data point from a Poisson distribution with mean $\alpha/N$, as proposed in~\cite[]{IBP}.

\subsection{Posterior Distribution over the Pseudo-observations}\label{subsec:sampleY}
In Algorithm~\ref{alg:inference}, we need to sample the pseudo-observations $y_{nr}^d$  and the auxiliary variables in $\Psi^d$ from their corresponding posterior distributions.  The posterior distributions for $y_{nr}^d$, and for $\Psi^d$, if needed, for all the considered types of data are given by:
\begin{enumerate}
\item For real-valued observation: 
\begin{equation}
p(y_{n1}^d|x^d_n,\zn,\Bb^d)
=\Ncal\left(y_{n1}^d\bigg|\left(\frac{\zn\bd_1}{\sigma^2_y}+\frac{f_{d}^{-1}(x^d_n)}{\sigma^2_u}\right)\left(\frac{1}{\sigma^2_y}+\frac{1}{\sigma^2_u}\right)^{-1},\left(\frac{1}{\sigma^2_y}+\frac{1}{\sigma^2_u}\right)^{-1}\right),
\end{equation}
where $f_{d}^{-1}: \Re \rightarrow \Re$.
\item For positive real-valued observations:
\begin{equation}
p(y_{n1}^d|x^d_n,\zn,\Bb^d)=\Ncal\left(y_{n1}^d\bigg|\left(\frac{\zn\bd_1}{\sigma^2_y}+\frac{f_{d}^{-1}(x^d_n)}{\sigma^2_u}\right)\left(\frac{1}{\sigma^2_y}+\frac{1}{\sigma^2_u}\right)^{-1},\left(\frac{1}{\sigma^2_y}+\frac{1}{\sigma^2_u}\right)^{-1}\right), 
\end{equation}
where $f_{d}^{-1}: \Re_+ \rightarrow \Re$.

\item  For categorical observations: 
\begin{equation}\label{eq:sampleY1}
\begin{split}
&p(y_{nr}^d | x_{n}^d=T, \zn ,\Bb^d) =  \left\lbrace \begin{array}{ll}
 \Ncal(y_{nr}^d| \zn \bd_r, \sigma^2_y)\mathbb{I}(y_{nr}^d>\max_{j\neq r}(y_{nj}^d)) & \mathrm{If} \quad r=T\\
 \Ncal(y_{nr}^d| \zn \bd_r, \sigma^2_y) \mathbb{I}(y_{nr}^d< y_{nT}^d) & \mathrm{If} \quad r\neq T
\end{array}\right.
\end{split}
\end{equation}
In Equation~\eqref{eq:sampleY1}, if $x^d_n=T =r$, we sample $y_{nr}^d$ from a Gaussian {{left-truncated}} by $\max_{j\neq r}(y_{nj}^d)$. Otherwise, we sample from a Gaussian {{right-truncated}} by $y_{nr}^d$ with $r=x^d_n$. Sampling the variables $y_{nr}^d$ corresponds to solving a multinomial probit regression problem. In order for the model to be identifiable, without loss of generality, we assume that the regression function $f_{R_d}(\zn)$ is identically zero. Therefore, we fix $b_{kR_d}^d=0$ for all $k$.


\item For ordinal observations:
\begin{equation}\label{eq:sampleY}
p(y_{n1}^d | x_{n}^d=r, \zn ,\Bb^d) \sim  \Ncal(y_{n1}^d| \zn \bd_1, \sigma^2_y) \mathbb{I}(\theta^d_{r-1} < y_{n1}^d \leq \theta^d_{r}). 
\end{equation}
In this case, we sample $y_{n1}^d$ form a Gaussian left-truncated by $\theta^d_{r-1}$ and right-truncated by $\theta^d_{r}$. 
In this case, we also need to sample the {{threshold values}} $\theta^d_r$ with $r=1,\ldots, R_d-1$ as
\begin{equation}\label{eq:sampleY}
\begin{split}
p(\theta_r^d | y_{n1}^d) \sim & \Ncal(\theta_r^d| 0, \sigma^2_{\theta})  \mathbb{I}(\theta_r^d > \max(\theta_{r-1}^d , \max_n(y_{n1}^d|x_{n}^d=r) )\\ & \times \mathbb{I}(\theta_r^d < \min(\theta_{r}^d , \min_n(y_{n1}^d|x_{n}^d=r+1)).
\end{split}
\end{equation}
In this case, {{sampling the variables $y_{n1}^d$ corresponds to solving}} an ordered probit regression problem, where the thresholds $\{\theta_r \}_{r=1}^{R_d}$ are unknown.
In order for the model to be identifiable we set one of the thresholds, $\theta_1$ in our case, to zero.

\item For count observations:
\begin{equation}\label{eq:sampleY}
p(y_{n1}^d| x_{n}^d, \zn ,\Bb^d) =  \Ncal(y_{n1}^d| \zn \bd_1, \sigma^2_y) \mathbb{I}(f_{\Re_+}^{-1}(x_n^d) \leq y_{n1}^d<f^{-1}(x_n^d+1)),
\end{equation}
where $f_{\Re_+}^{-1}: \Re_+ \rightarrow \Re$. 
{{We sample}} $y_{n1}^d$ from a Gaussian {{left-truncated }} by $f_{\Re_+}^{-1}(x_n^d)$ and {{right-truncated}} by $f_{\Re_+}^{-1}(x_n^d+1)$.
\end{enumerate}

\section{Applications}
\label{sec:applications}
In this section, we apply the proposed model to solve two different tasks on several real-world datasets. In Section~\ref{subsec:tableCompletion}, we focus on a prediction task in which we aim to estimate and replace the missing data, which is assumed to be missing completely at random. These results have been previously introduced in \cite[]{Valera_NIPS2015}. 
Then, in Section~\ref{subsec:data_exploration}, we focus on a data analysis task on several real-world datasets from different application domains including psychiatry, clinical trials and politics. Weexploratory data analysis show how to use the proposed model to perform exploratory data analysis, i.e., to find the latent structure in the data and capture the statistical dependencies among the objects and their attributes in the data. 

\subsection{Missing Data Estimation}\label{subsec:tableCompletion}
In this section, we use the proposed model to estimate missing data in heterogeneous databases, where we assume that {{the data is}} missing completely at random (MCAR). 
Missing data may occur in diverse applications due to different reasons. 
For example, 
participants of a survey may decide not to respond (or skip) some questions of the survey; 
participants in a clinical study may drop out during the course of the study; 
or users of a recommendation system can only rate a small fraction of the available books, movies, or songs, due to time constraints. 
The presence of missing values might be challenging when the data is used for reporting, information sharing and decision support, and as a consequence, handling missing data has captured the attention in diverse areas of data science such as machine learning, data mining, and data warehousing and management \cite[]{schafer2002missing,mazumder2010spectral}.

The extensive literature in probabilistic missing data estimation and imputation  focus on homogeneous databases which contain only either continuous data, usually modeled as Gaussian variables \cite[]{Todeschini2013}, or discrete data, that can be either modeled by discrete likelihoods \cite[]{Li2009} or simply treated as Gaussian variables \cite[]{BPMF_2008, Todeschini2013}.  
However, up to our knowledge, there is a lack of approaches for missing data estimation in heterogeneous databases, where the standard approach is to treat all the attributes, either continuous or discrete, as  Gaussian variables.

\xhdr{Experimental Setup} We  evaluate the predictive power of the proposed model at estimating missing data on five real databases, which are summarized in Table~\ref{table:datasets}. The datasets contain different numbers of instances and attributes, which cover all the discrete and continuous variables described in Section~\ref{sec:model}. 
According to our proposed model, the probability distribution over the observation matrix is fully characterized by the latent matrices $\Zb$ and $\{\Bb^d\}_{d=1}^D$  (as well as the auxiliary variables $\Psi^d$). Hence, if we assume the latent vector $\zn$ for the $n$-th datapoint and the weight factors $\Bb^d$ (and the auxiliary variables $\Psi^d$) to be known, we have a probability distribution over missing observations $\x_n^d$ from which we can obtain estimates for $\x_n^d$ by sampling from this distribution\footnote{Note that  sampling from this distribution might be computationally expensive. In this case, we can easily obtain samples of $\x_n^d$ by exploiting the structure of our model. In particular, we can simply sample the auxiliary Gaussian variables $y_n^d$ given $z_n$ and $\Bb^d$, and then obtain an estimate for $\x_n^d$ by applying the corresponding transformation, detailed in Section~\ref{sec:likFunctions}.}, or by simply taking either its mean, mode or median value, once the latent matrix $\Zb$ and the latent weight factors $\Bb^d$ (and $\Psi^d$) have been learnt. 

Here, we compare in terms of predictive log-likelihood the following methods for missing data estimation:
\begin{compactitem}
\item The proposed general table completion approach denoted by GLFM (see Section~\ref{sec:model}).
\item The standard linear-Gaussian IBP \cite[]{IBP} denoted by SIBP, treating all the attributes as Gaussian.
\item The Bayesian probabilistic matrix factorization approach \cite[]{BPMF_2008} denoted by BPMF, that also treats all the attributes in $\Xb$ as Gaussian distributed.
\end{compactitem}
For the GLFM, we consider for the real positive and the count data the following transformation that maps from the real numbers to the real positive numbers, $f(x)=\log(\exp(w x)+1)$, where $w$ is a user-defined hyper-parameter that scales the data. Before running the SIBP and the BPMF methods, we normalized each column in the matrix $\Xb$ to have zero-mean and unit-variance. 
This normalization ensures that the Gaussian likelihood evaluations of all the attributes describing the objects in each dataset are comparable, regardless of their discrete or continuous nature. As a consequence, ittherefore provides more accurate and fair results than applying the SIBP and BPMF directly on the data without prior normalization. 
%
Additionally, since both the SIBP and the BPMF assume continuous observations, when dealing with discrete data, we estimate each missing value as the closest integer value to the ``de-normalized'' Gaussian variable.

\begin{table}[t]
\small
\centering
\renewcommand{\tabcolsep}{4pt}
\begin{tabular}{|p{3.8cm}|p{1cm}|p{2.5cm}|p{7cm}|} \hline
Dataset & N & D & Description \\ \hline\hline
Statlog German credit dataset  \cite[]{German} & 1{,}000 & 20  (10 C + 4 O + 6 N)  &  Information about the credit risks of the applicants.\\ \hline
 QSAR biodegradation dataset \cite[]{BioDeg}& 1{,}055 & 41 (2 R + 17 P + 4 C + 18 N) &  Molecular descriptors of biodegradable and non-biodegradable chemicals. \\ \hline
Internet usage survey dataset \cite[]{Internet}& 1{,}006 & 32 (23 C + 8 O + 1 N)  &   Responses of the participants to a survey related to the usage of internet. \\ \hline
Wine quality dataset \cite[]{Wine}& 6{,}497 & 12 (11 P + 1 N) & Results of physicochemical tests realized to different wines.\\ \hline
NESARC dataset \cite[]{PsiquiatrasJMLR} & 43{,}000 & 55 C &   Responses of the participants to a survey related to personality disorders. \\ \hline
\end{tabular}
\vspace*{-5pt}
\caption{\textbf{Description of datasets.} `R' stands for real-valued variables, `P' for positive real-valued variables, `C' for categorical variables, `O' for ordinal variables and `N' for count variables}\label{table:datasets}
\vspace*{-10pt}
\end{table}

\begin{figure}[t]
\centering
\subfloat[Statlog.]
{\includegraphics[width=0.329\textwidth]{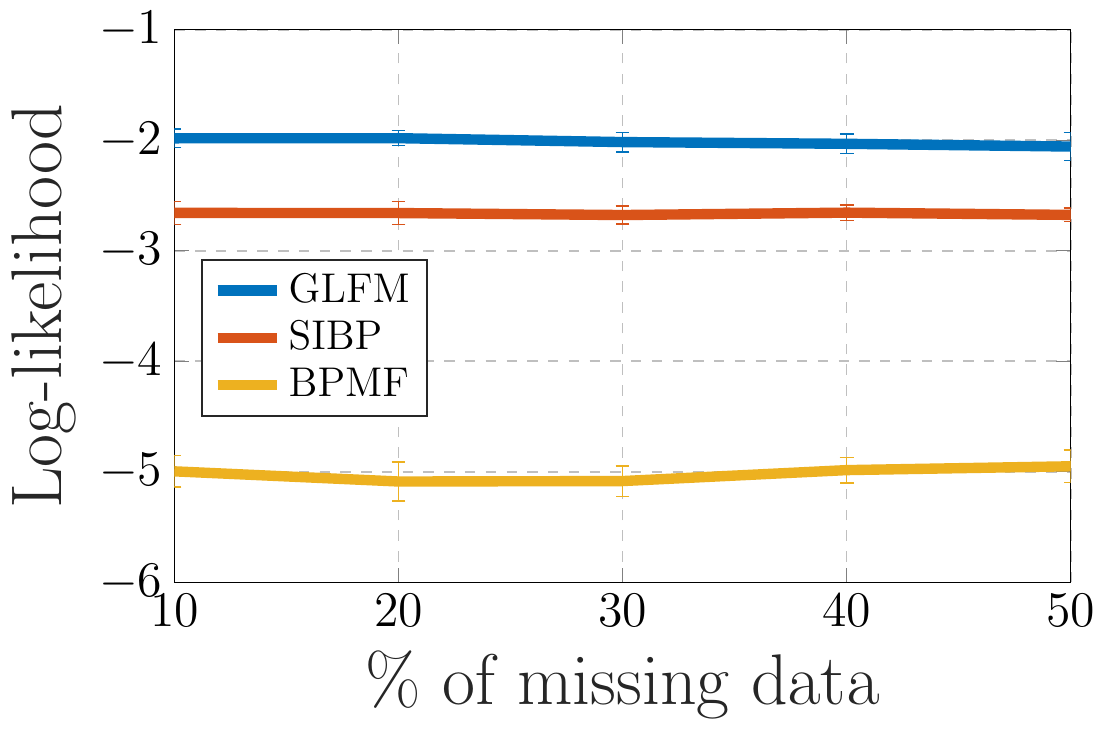}\label{fig:German}}
\subfloat[QSAR biodegradation.]
{\includegraphics[width=0.329\textwidth]{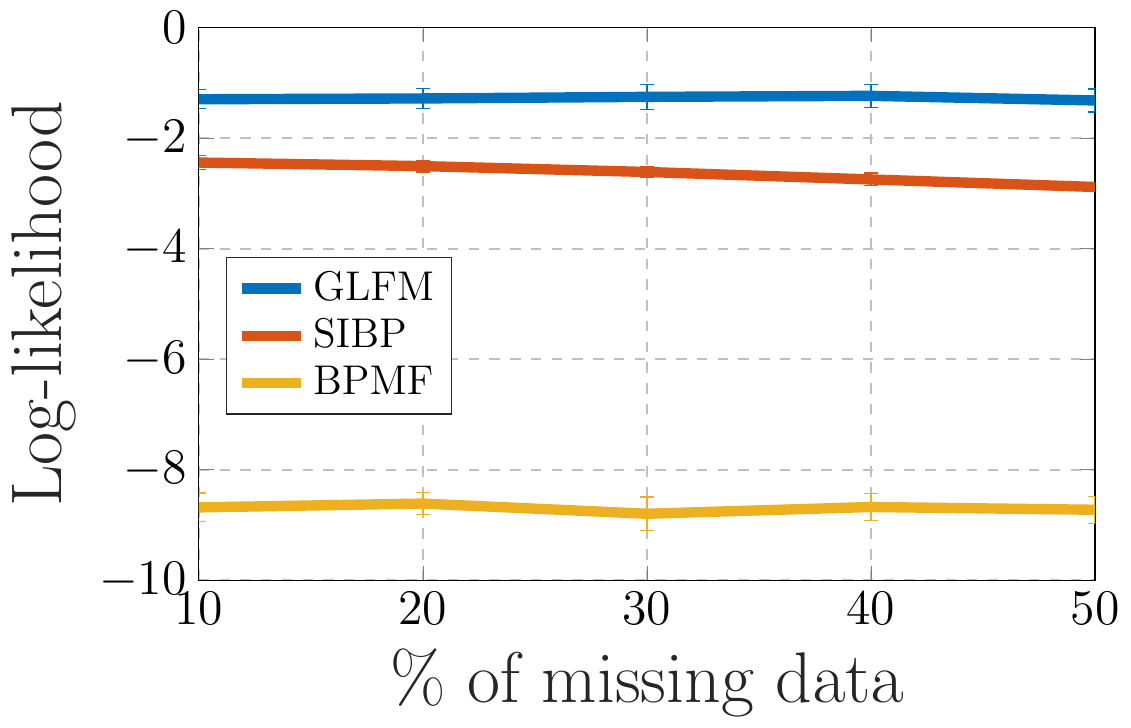}\label{fig:BioDeg}}
\subfloat[Internet usage survey.]
{\includegraphics[width=0.329\textwidth]{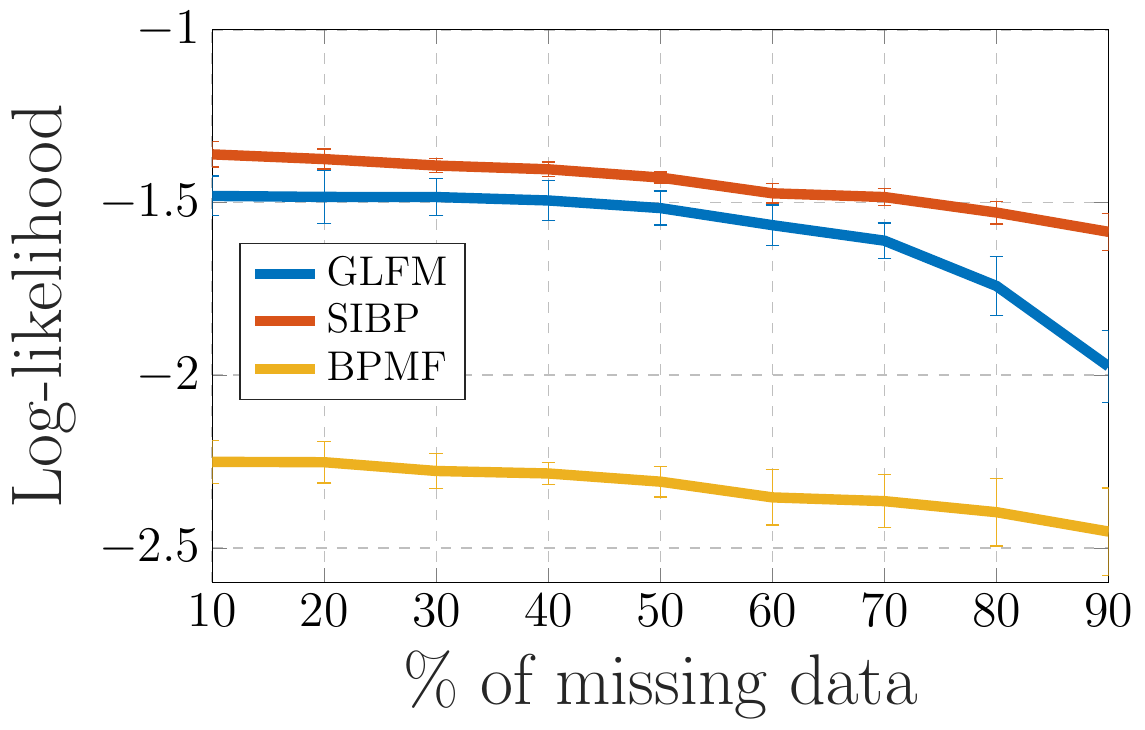}\label{fig:German}}
\vspace*{-10pt}
\subfloat[Wine quality.]
{\includegraphics[width=0.329\textwidth]{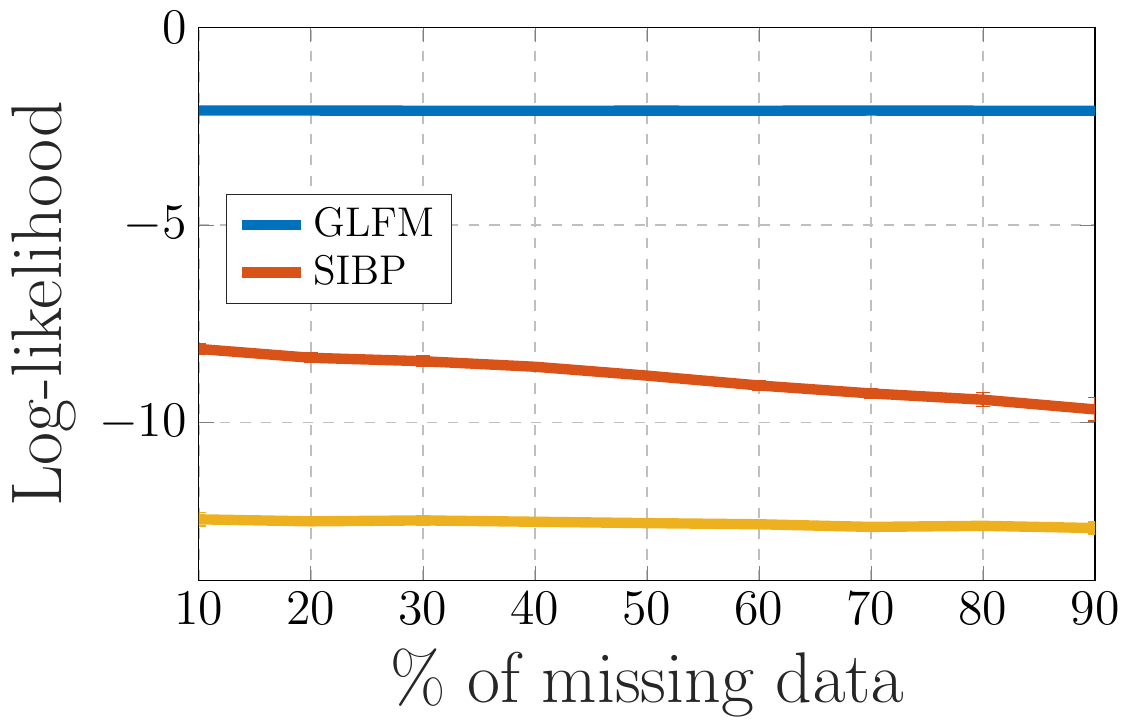}\label{fig:Wine}}
\subfloat[Nesarc database.]
{\includegraphics[width=0.329\textwidth]{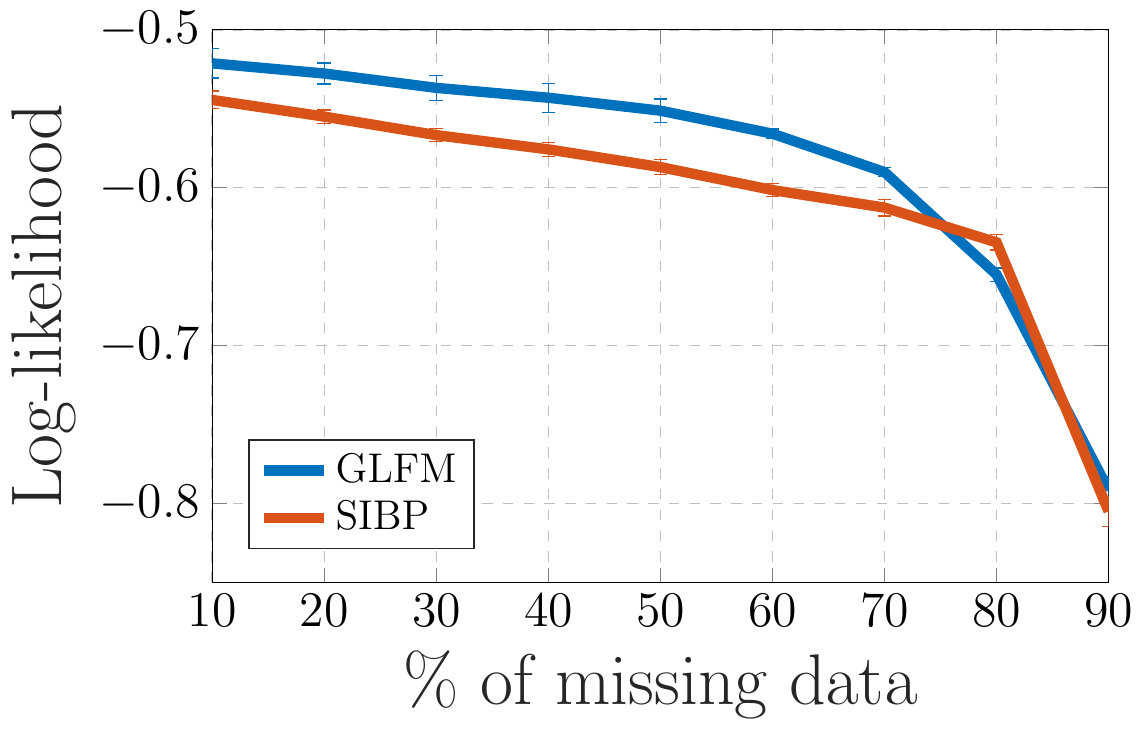}\label{fig:Nesarc}}
\caption{\textbf{Average test log-likelihood per missing datum versus percentage of missing data.} The `whiskers' show one standard deviation away from the average test log-likelihood. 
}
\label{fig:percentage}
\vspace*{-10pt}
\end{figure}

\begin{figure}
\vspace*{-10pt}
\centering
\subfloat[Statlog.]
{\includegraphics[width=1\textwidth]{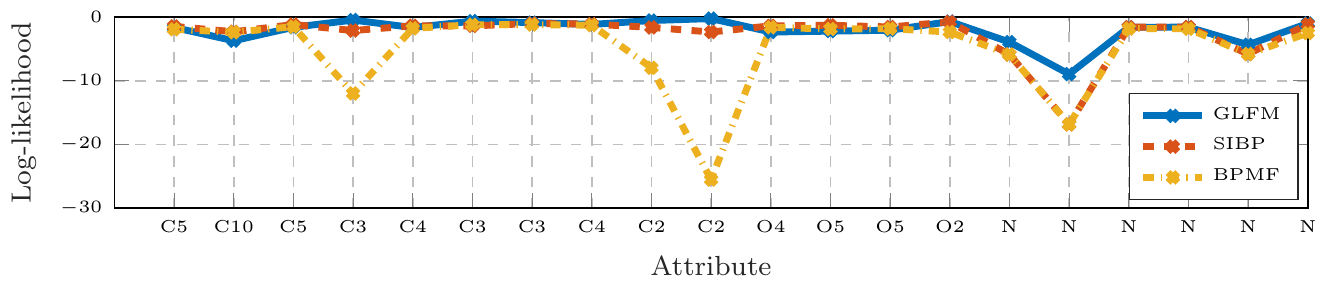}\label{fig:German_D}} \\
\vspace*{-10pt}
\subfloat[QSAR biodegradation.]
{\includegraphics[width=1\textwidth]{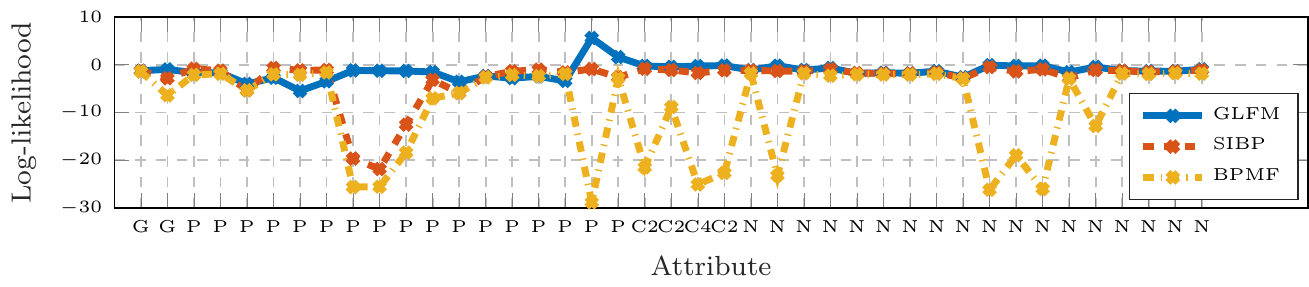}\label{fig:BioDeg_D}} \\
\vspace*{-10pt}
\subfloat[Internet usage survey.]
{\includegraphics[width=1\textwidth]{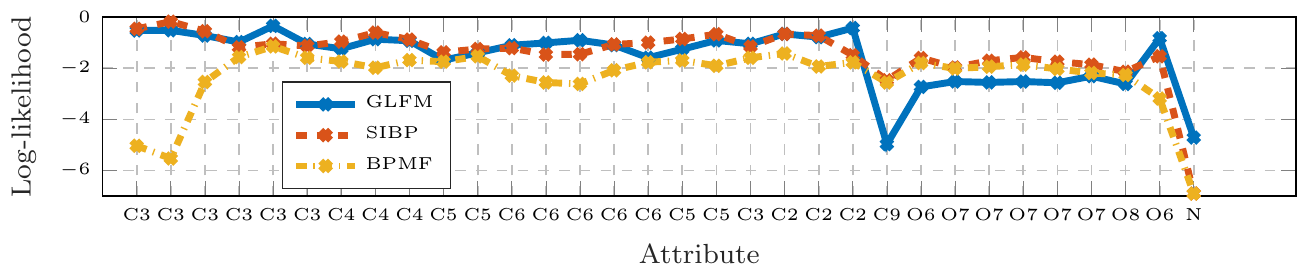}\label{fig:Internet_D}} \\
\vspace*{-10pt}
\subfloat[Wine quality.]
{\includegraphics[width=1\textwidth]{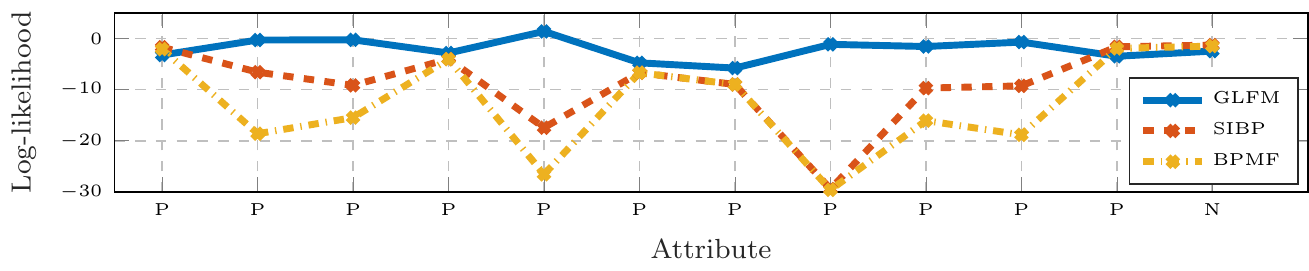}\label{fig:Wine_D}} \\
\vspace*{-10pt}
\subfloat[Nesarc database.]
{\includegraphics[width=1\textwidth]{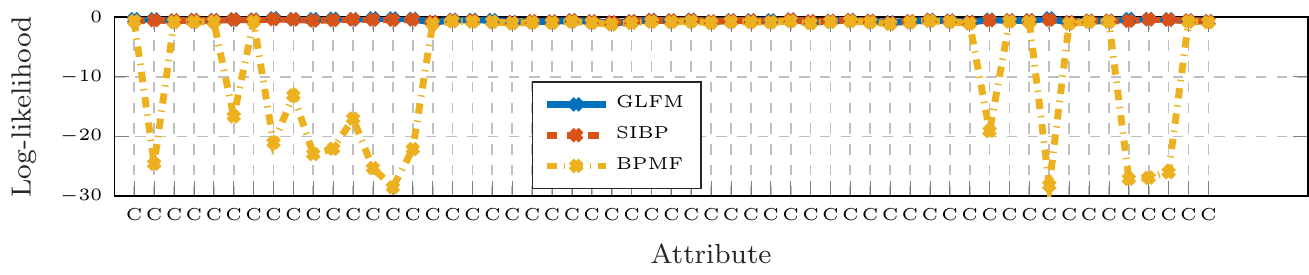}\label{fig:Nesarc_D}} \\
\vspace*{-10pt}
\caption{\textbf{Average test log-likelihood per missing datum in each dimension.} Here, we consider 50\% of missing data. In the x-axis `R' stands for real-valued variables, `P' for positive real-valued variables, `C' for categorical variables, `O' for ordinal variables and `N' for count variables. The number that accompanies `C' or `O' corresponds to the number of categories. In the Nesarc dataset, all the variables are binary, i.e., `C2'.}
\vspace*{-10pt}
\label{fig:TLK_D}
\end{figure}

%
%
\xhdr{Results} Figure~\ref{fig:percentage} plots the average predictive log-likelihood per missing value as a function of the percentage of missing data. Each value in Figure~\ref{fig:percentage} was obtained by averaging the results across 20 independent split sets where the missing values were randomly chosen. 
In Figures~\ref{fig:German} and \ref{fig:BioDeg}, we cut the plot {{at a missing percentage of 50\%}} because, in these two databases, the discrete attributes present a mode value that appears for more than 80\% of the instances. As a consequence, the SIBP and the BPMF algorithms assign probability close to one to the mode, which results in an artificial increase in the average test log-likelihood for larger percentages of missing data. 
For BPMF, we have used different numbers of latent features (in particular, 10, 20 and 50), although we only show the best results for each database, specifically, $K=10$ for the NESARC and the wine databases, and $K=50$ for the remainder. Both the GLFM and the SIBP have not learnt a number of binary latent features above 25 in any case. In Figure~\ref{fig:Nesarc}, we only plot the test log-likelihood for the GLFM and SIBP because BPMF provides much lower values. 
%
%
As expected, we observe in Figure~\ref{fig:percentage} that the average test log-likelihood decreases for the three models as the number of missing values increases (flat shape of the curves are due to the logarithmic scale of the y-axis). In this figure, we also observe that the proposed GLFM outperforms SIBP and BPMF for four of the databases, being SIBP slightly better for the Internet database. The BPMF model presents the worst test log-likelihood in all databases. 

Successively, we analyzed the performance of the three models for each kind of discrete and continuous variables. Figure~\ref{fig:TLK_D} shows the average predictive likelihood per missing value for each attribute in the table, which corresponds to each dimension in $\Xb$. In this figure, we have grouped the dimensions according to the kind of data that they contain, the x-axis shows the number of considered categories for the case of categorical and ordinal data (in the case of the NESARC database, all the attributes are binary). The figure shows that the GLFM presents similar performance for all the attributes in the five databases, while  for the SIBP and the BPMF models, the test log-likelihood falls drastically for some of the attributes, with this effect being more dramatic in the case of  BPMF (it explains the low log-likelihood in Figure~\ref{fig:percentage}). This effect is even more evident in Figures~\ref{fig:BioDeg} and \ref{fig:Wine}. 
%
In Figures~\ref{fig:percentage} and \ref{fig:TLK_D}, we observe that both IBP-based approaches (GLFM and SIBP) outperform BPMF, with the proposed GLFM being the one that best performs across all databases.  We can conclude that, unlike  BPMF and  SIBP,  GLFM provides accurate estimates for the missing data regardless of their discrete or continuous nature.

\subsection{Data Exploration}\label{subsec:data_exploration}
In this section, we describe how to use the GLFM for a data exploration task: the latent structure that underlies the data is discovered and analized in three different application domains. Specifically, we  use the proposed GLFM in the context of i) clinical trials to discover the effects of a new drug for prostate cancer; ii) psychiatry to capture the impact of social background in the development of mental disorders;  and iii) politics to identify meaningful demographic profiles, together with their geographic location, and voting tendencies in the United States. 

The main goal of this section is to show how to include the specific domain knowledge into the proposed GLFM  to ease the data exploration process.  
In particular, we here show examples of how to select the input data for the GLFM, as well as how to input these data into the model, in order to obtain interpretable results that can be used  to get a better understanding of the corresponding knowledge domain.

\subsubsection{Drug effect in a clinical trial for prostate cancer}
Clinical trials are conducted to collect data in order to determine the safety and efficacy of a new drug before it can be sold in the consumer market. 
Concretely, the main goal of clinical trials is to prove the efficacy of a new treatment for a disease while ensuring its safety, i.e., check whether its adverse effects 
remain low enough for any dosage level of the drug. 
As an example, the publicly available \textit{Prostate Cancer dataset}\footnote{Database available at: \url{http://biostat.mc.vanderbilt.edu/wiki/Main/DataSets}}  collects data of a clinical trial that analized the effects of  the drug diethylstilbestrol (DES) as a treatment against prostate cancer. 
The dataset contains information about 502 patients with prostate cancer in stages\footnote{The stage of a cancer describes the size of a cancer and how far it has grown. Stage 3 means that the cancer is already quite large and may have started to spread into surrounding tissues or local lymph nodes. Stage 4 is more severe, and refers to a cancer that has already spread from where it started to another body organ. This is also called secondary or metastatic cancer. Find more details in~{\url{http://www.cancerresearchuk.org/about-cancer/what-is-cancer/stages-of-cancer}}} 3 and 4, who entered a clinical trial during 1967-1969 and were randomly allocated to different levels of treatment with DES.
The prostate cancer dataset have been used by several studies~\cite[]{byar1980,Kay1986, Lunn1995} to analyze the survival times of the patients in the clinical trial and the causes behind their death. 
These studies have pointed out that a large dose of the treatment tend to reduce the risk of a cancer death, but it might also result in an increased risk of cardiovascular death.
In this section, we apply the proposed GLFM to the {Prostate Cancer dataset} to show that the proposed model can be efficiently used to discover  the statistical dependencies in the data, which in this example corresponds to the effect of the different levels of treatment with DES in the suffering of prostate cancer and cardiovascular diseases. 

\xhdr{Experimental Setup}
The  prostate cancer dataset consists of 502 patients and 16 attributes, from which we make use of the five attributes listed in Table~\ref{tab:prostate}. 
The selection of these five attributes allows us to focus onlytherefore on capturing the statistical dependencies between the target attributes, i.e, the relationship between the different levels of treatment with DES and the suffering of prostate cancer and cardiovascular diseases.
In our experiments, we sample the variance of the pseudo-observations in each dimension and choose $\alpha=5$, $\sigma_B^2=1$, and  $\sigma_{\theta}^2=1$. We also consider for the positive real  and count data the following transformation that maps from the real numbers to the positive real numbers: $f(x) = \log(w\cdot(x - \mu) + 1)$, where $\mu = \min(\mathbf{x}^d)$ and $w=2/\mathrm{std}(\mathbf{x}^d)$ are data-driven parameters whose objective is to shift and scale the data. 
In order to obtain more interpretable results, we also activate the bias term, i.e., a latent feature forced to be active for all patients. 
\begin{table}[t]
\centering
\small{\begin{tabular}{ |p{6.5cm}|l|} \hline
Attribute description & Type of variable \\ \hline \hline 
Stage of the cancer & Categorical with 2 categories \\ 
DES treatment level & Ordinal with 3 categories  \\ 
Tumor size in cm$^2$ & Count data  \\ 
Serum Prostatic Acid Phosphatase (PAP)& Positive real-valued \\
Prognosis Status (outcome of the disease) & Categorical with 4 categories \\ 
 \hline
\end{tabular}}
\caption{\textbf{List of considered attributes for the Prostate Cancer dataset.}}
\label{tab:prostate}
\end{table}

 \begin{table}[th]
 \vspace*{-3mm}
\small
\centering
\begin{tabular}{|c|c| p{9.7cm} |} \hline
Feature & Empirical Prob. & Main implications \\ \hline\hline
Feature F1 & $0.1952$  &{Favors stage 3, low DES levels and prostatic death} \\ \hline
Feature F2 & $0.2689$  & {Favors stage 3, highest DES levels, and cardiovascular death}\\ \hline
Feature F3 & $0.1594$ & {Favors stage 4, low DES levels, and mid-level prostatic death}  \\  \hline
Feature F4 & $0.1155$ &  {Favors stage 4, low DES levels, and most severe prostatic cancer} \\ \hline
\end{tabular}
\caption{\textbf{Empirical feature activation probabilities in the Prostate Cancer dataset.} We show the empirical probability of possessing at least one latent feature. These probabilities are directly computed from the inferred IBP matrix $\Zb$. Additionally, the table summarizes the main implications of the activation of each latent feature. }\label{tab:probsProstate}
\vspace*{-3mm}
\end{table}
 \begin{table}[h]
\small
\centering
\begin{tabular}{|c|c|c|c|c|c|c|c|c|c|} \hline
Patterns & (0000) & (0100) & (1000) & (0010) & (0001) & (1100) & (0110) & (0101) & (1010) \\ \hline\hline
Empirical Prob. & $0.4641$ & $0.1394$ & $0.0936$ & $0.0757$  & $0.0518$ & $0.0438$ & $0.0359$ &  $0.0259$ &  $0.0219$ \\ \hline
\end{tabular}
\caption{\textbf{Empirical probability of pattern activation for the top-nine most popular patterns.} These probabilities are computed directly from the inferred IBP matrix $\Zb$.}\label{tab:patternsProstate}
\vspace*{-3mm}
\end{table}

\begin{figure}[t]
	\centering
	\subfloat[Type of Cancer]	
	{\includegraphics[width=0.45\textwidth]{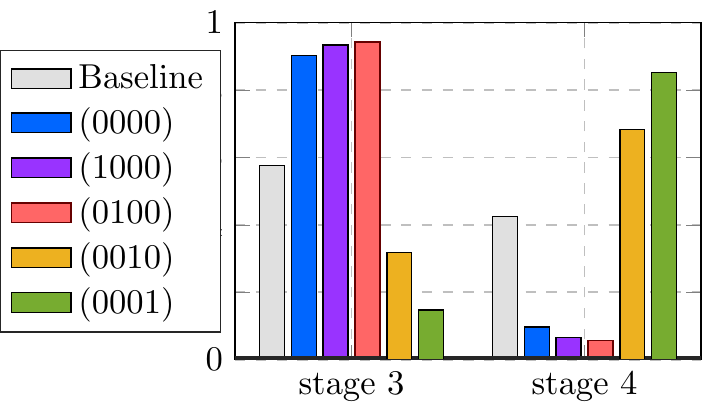}\label{fig:stafe}}\quad\quad
    \subfloat[Drug Level]
    {\includegraphics[width=0.33\textwidth]{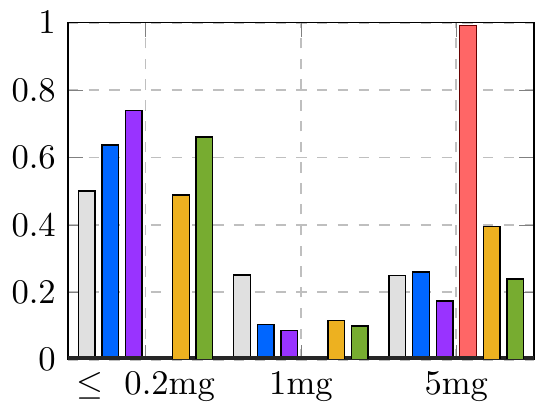}\label{fig:drug}}
    
    \vspace{-0.2cm}
\begin{tabular}{p{0.5\textwidth} p{0.5\textwidth}}
  \vspace{0pt} \subfloat[Size of Primary Tumor (cm$^2$)]
  {\includegraphics[width=0.5\textwidth]{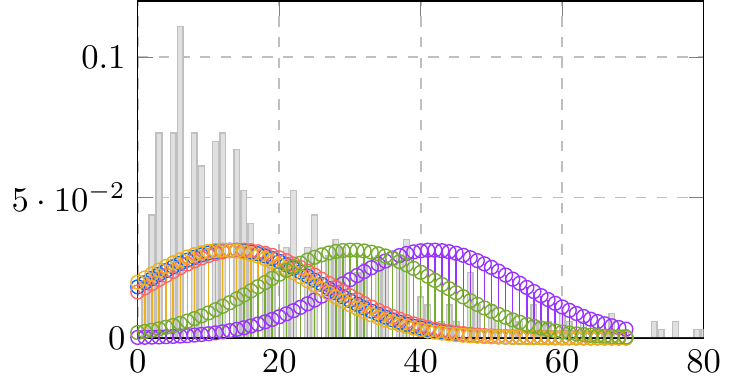}\label{fig:tumor}}  
  &
  \vspace{-1pt} \subfloat[Serum Prostatic Acid Phosphatase]
    {\includegraphics[width=0.45\textwidth]{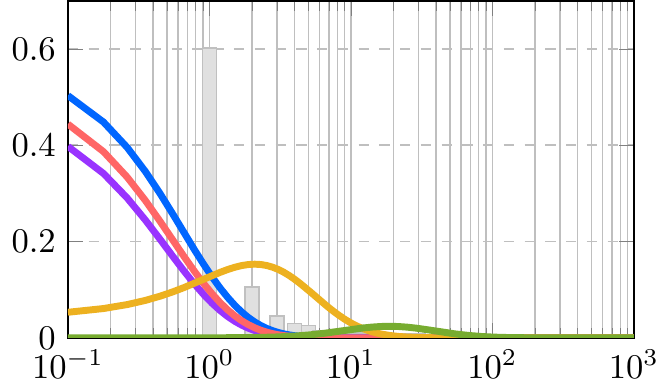}} 
  \label{fig:prosta4}
\end{tabular}

 \vspace{-0.1cm}
\subfloat[Prognosis Status]
  {\includegraphics[width=1\textwidth]{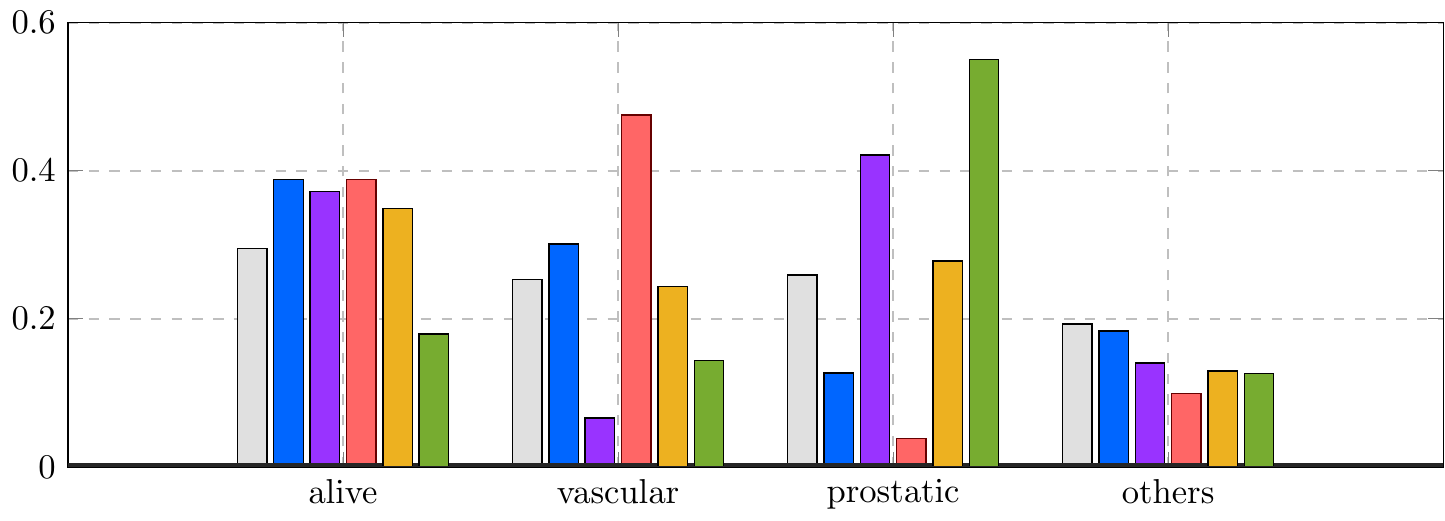}}\label{fig:prosta5}
     \vspace{-0.2cm}
	\caption{\textbf{Data exploration of a prostate cancer clinical trial.} We depict the effect of each latent feature on each attribute. Panels (a)-(d) shows different indicators of the prostate cancer, as well as the dose level of DES. Panel (d) corresponds to Prognosis Status, which indicates whether the patient either is alive or dies from one of the following three causes: vascular disease, prostatic cancer, or other reason. The baseline refers to the empirical distribution of each attribute in the whole dataset.}\label{fig:prostate}
\end{figure}


\xhdr{Results}
After running our model, we obtain four latent features,  whose empirical probabilities of activation and main implications (which are discussed further below) are shown in Table~\ref{tab:probsProstate}. 
%
Additionally, Table~\ref{tab:patternsProstate} shows the nine most common latent feature vectors, referred here as \emph{feature patterns}, which capture over 95\% of the observations. 
In order to study the effect of the latent features on each attribute of the dataset, Figure~\ref{fig:prostate} shows the inferred distribution of each attribute for the five most common patterns, which happen to only have one active latent feature (plus the bias term). 
We remark that GLFM assumes that the contribution of each latent feature is additive, therefore, patterns with more than one active latent feature can be seen as combinations of the depicted patterns.\footnote{In the case of patterns with multiple active latent features, the bias term should be counted only once.}

In Figure~\ref{fig:prostate}, we can distinguish two groups of features. 
The first group accounts for patients in stage 3 and includes the bias term and the two first latent features.
 Within this group,  the bias term -- or equivalently pattern (0000) -- and the first feature --or equivalently pattern (1000) -- account for patients in stage 3 with a low average level of treatment with DES, as shown in Figure~\ref{fig:drug}.
 However, while the bias term models patients with low probability ($\sim15\%$) of prostate cancer death, the first feature accounts for patients with higher probability ($\sim40\%$) of prostate cancer death, which can be explained by a larger tumor size, as shown in Figure~\ref{fig:tumor}. 
The second feature -- or equivalently pattern (0100) --captures patients who exclusively received a high dosage (5~mg) of the drug, as shown in Figure~\ref{fig:drug}. 
These patients present a small tumor size and the lowest probability of prostatic cancer death, suggesting a positive effect of the drug as treatment for the cancer. However, they also present a significant increase in the  probability  of dying from a vascular disease ($\sim50\%$), indicating a potential adverse-effect of the drug that increases the risk of suffering from cardio-vascular diseases. Such observation is in agreement with previous studies~\cite[]{byar1980, Kay1986, Lunn1995}. 

The second group of features corresponds to the activation patterns (0010) and (0001), and accounts for patients in stage 4 with mild and severe conditions, respectively.
 In particular, the third feature  corresponds to patients with small tumor size, but intermediate values for the PAP biomarker, suggesting a certain spread in the degree of the tumor compared to the features in the first group, but not as severe as for patients with pattern (0001). Indeed, pattern (0001) models those patients in stage 4 with relatively high tumor size and the highest PAP values -- it is thus not surprising that those patients present the highest probability (above 50\%) of prostatic death.
%
 
 %


\subsubsection{Impact of Social Background on Mental Disorders}
In this section, we aim at extending the analysis in~\cite[]{PsiquiatrasJMLR} to account for  the influence of the social background of subjects (such as age, gender, etc.) in the probability of a subject suffering from a comorbid disorder. 
To this end, in addition to the diagnoses of the the 20 most common psychiatric disorders detailed below, we  also make use of the information provided by the \textit{NESARC database}\footnote{Database available at:
\url{http://aspe.hhs.gov/hsp/06/catalog-ai-an-na/nesarc.htm}}, which includes information both on the mental condition and on the social background of participants. 

Several studies have analyzed the impact of social background in the development of mental disorders. These studies usually focus on the relation between a mental disorder and a specific aspect of the social background of subjects. Some examples in this area study the relation between depression and gender~\cite[]{Weissman1993, Kessler199385}, or the link between common mental disorders and poverty or social class~\cite[]{Weich1998, Dohrenwend75,Hollingshead1953}. 
%
%
%
Other studies~\cite[]{Blanco2012, PsiquiatrasJMLR} have focused on finding and analyzing the co-occurring (comorbidity) pattern among the 20 most common psychiatric illnessesto our knowledge. 
These studies found that the 20 most common disorders can be divided into three meta-groups of disorders:  i) externalizing disorders, which include substance use disorders (alcohol abuse and dependence, drug abuse and dependence and nicotine dependence); ii)  internalizing disorders, which include mood and anxiety disorders (major depressive disorder (MDD), bipolar disorder and dysthymia, panic disorder, social anxiety disorder (SAD), specific phobia and generalized anxiety disorder (GAD), and pathological gambling (PG));  and iii) personality disorders (avoidant, dependent, obsessive-compulsive (OC), paranoid, schizoid, histrionic and antisocial personality disorders (PDs)). 
Additionally, they also found that comorbid or co-occurring disorders tend to belong to the same group of disorders~\cite[]{valera2016infinite}.
However, to our knowledge, there is a lack of work in the study of the impact of social background in the suffering of comorbid disorders.

%

\xhdr{Experimental Setup}
The NESARC database contains the responses of a representative sample of the U.S.
population to a survey with questions related to the social background of participants, alcohol
and other drug consumption, and behaviors related to mental disorders. The first wave of NESARC
sampled the adult U.S. population with over 43,000 respondents who answered almost
3,000 questions. The dataset also include the diagnoses for each of the participants of the survey. 
In this experiment,  in addition to the diagnoses of the 20 most common psychiatric disorders described above, we include one by one each of the social background questions as input data to the proposed model. Table~\ref{tab:20Q+} summarizes the considered questions and how we introduce them into our model as input variables. 
Here, we introduce each attribute related to the social background of the participants independently to ensure that the model captures the dependencies between latent disorders and social background, instead of correlations among the different aspect of the social background. 
Note also that the diagnoses of the 20 psychiatric disorders correspond to categorical variables with two possible categories, e.g., a patient suffering or not from a disorder.

For the following experimental results, {{we run the inference algorithm in Section~\ref{sec:inference} for each question independently }} with $\alpha=5$, $\sigma_B^2=1$, $\sigma_y^2=1$, $\sigma_{\theta}^2=1$, and consider for the positive real  and count data the following transformation that maps from the real numbers to the positive real  numbers: $f(x)=x^2$, which we chose to show that the proposed model works with any differentiable and invertible function. 
Similarly to~\cite[]{PsiquiatrasJMLR}, we activate the bias term, i.e., an additional latent feature which is active to all the subjects in the data set,  so that we do not sample the rows of $\Zb$ corresponding to those subjects who do not suffer from any of the $20$ disorders, but instead fix their latent features to zero. The idea is that the bias captures the population that does not suffer from any disorder, while the rest of active features in matrix $\Zb$ characterize the disorders.  As we will see next, the use of the bias term aims to ease the interpretability of the inferred latent features. 

\begin{table}[t]
\centering
\small{\begin{tabular}{|p{6.5cm}|l|} \hline
Attribute description & Type of variable\\ \hline \hline
Gender & Categorical with 2 categories\\ \hline
Age & Count data\\ \hline
Census region & Categorical with $4 $ categories\\ \hline
Race/ethnicity &  Categorical with $5$ categories\\ \hline
Marital status & Categorical with $6$ categories\\ \hline
Highest grade or years of school completed & Ordinal with $14$ categories\\ \hline
\end{tabular}}
\caption{\textbf{List of considered social background attributes.} We look for correlations between each of these attributes with the twenty most common psychiatric disorders among the subjects in the NESARC database.}
\label{tab:20Q+}
\vspace{-2mm}
\end{table}

\xhdr{Results}
After running our model, we find that the census region, race/ethnicity, marital status and educational level (i.e., highest grade or years of school completed) do not appear to have any influence in the comorbidity patterns of the 20 most common psychiatric disorders. In contrast, as detailed below,  gender and age 
of the participants influence the probability of suffering from a set of comorbid (co-occurring) disorders. 

\begin{figure}[]
\vspace*{-5pt}
\centering
\begin{tabular}{c}
\subfloat[Probability of suffering from each disorder]
{\includegraphics[width=1\textwidth]{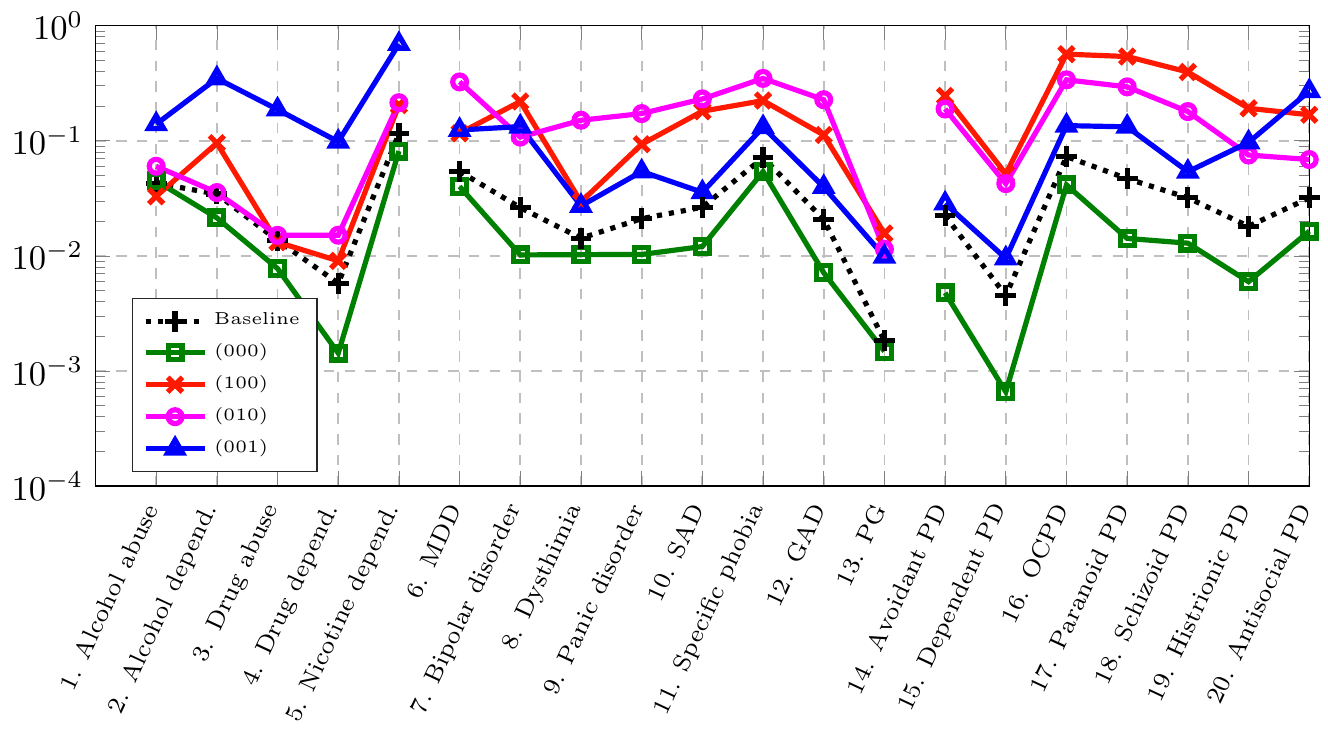}\label{fig:20Q+_5}}\\ \vspace*{-3mm}
\subfloat[Gender]
{\includegraphics[width=.7\textwidth]{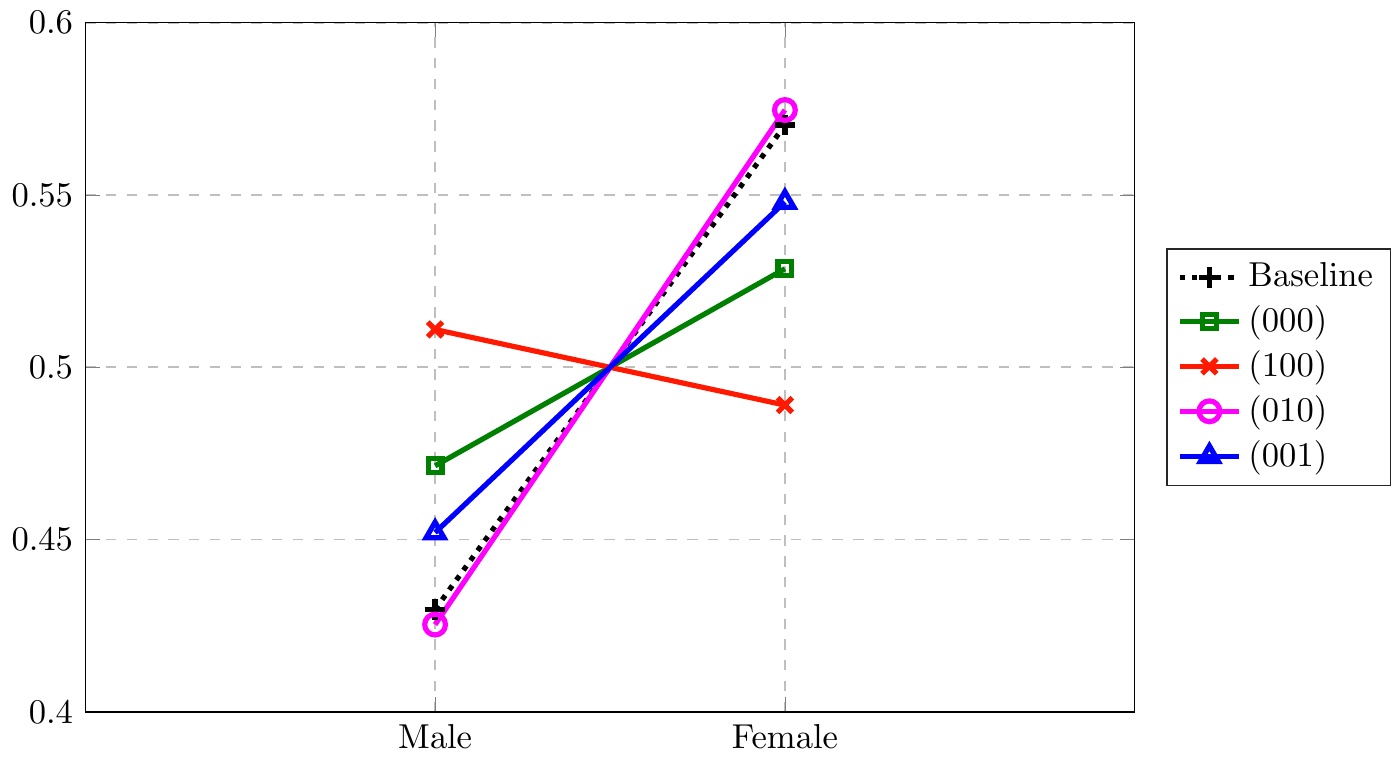} \label{fig:21Q_5}} 
\end{tabular}
\caption{\textbf{Feature effects including gender in the analysis.} (a) Probabilities of suffering from the $20$ considered disorders and (b) probability of being male and female for the latent feature vectors $\zn$ shown in the legend and for the baseline.}
\label{fig:gender}
\end{figure}

\xhdr{Gender}
We model the gender information of the participants in the NESARC as a categorical variable with two categories: $\{$`male', `female'$\}$.  The percentage of males in the NESARC is approximately $ 43\%$. 
%
In this case, the GLFM found three latent features, whose empirical probabilities can be found in Table~\ref{table:probsSex}. 
%
Moreover, Table~\ref{tab:patternsGender} shows empirical probability of each feature pattern in the dataset. Here, we observe that the three latent feature activate mostly in isolation, being the combination of two features rare (below ~1\%). 
Figure~\ref{fig:20Q+_5} shows the probability of meeting each diagnostic criteria for the latent feature vectors $\zn$ listed in the legend and in the database (baseline). Note that the obtained latent features are similar to the ones in \cite[]{PsiquiatrasJMLR}, i.e., Feature F1 -- pattern $(100)$ -- mainly models the seven personality disorders (PDs), Feature F2, which corresponds to pattern $(010)$, models the alcohol and drug abuse disorders and the antisocial PD, while Feature F3 -- pattern $(001)$ -- models the anxiety and mood disorders. Additionally, in Figure~\ref{fig:21Q_5}, we show the probability of being male and female for the latent feature vectors $\zn$ shown in the legend and the empirical probability of being male and female in the database (baseline).

In Figure~\ref{fig:21Q_5}, we observe that having no active features (pattern $(000)$, which captures people that do not suffer from any disorder), increases the probability of being male with respect to the baseline probability, therefore, it indicates that females tend to suffer in a higher extent from psychiatric disorders. 
Additionally, we observe that  Feature F1 -- pattern $(100)$ -- increases the probability of being male, while Feature F3 -- pattern $(001)$ --  increases the probability of being female. Hence, from the analysis of Figure~\ref{fig:21Q_5}, we can conclude that, while women suffer more frequently from mood and anxiety disorders than men, PDs are more common in men.

\begin{table}[h]
\small
\centering
\renewcommand{\tabcolsep}{4pt}
\begin{tabular}{|c|c|p{10cm}|} \hline
Feature & Empirical Prob. & Main implications \\ \hline\hline
Feature F1 & $ 0.0341$  & Increases the probability of personality disorders and male gender \\
Feature F2 & $0.0470$  & Increases the probability of alcohol and drug abuse disorders  \\
Feature F3 & $0.0460$ & Increases the probability of anxiety and mood disorders and female gender\\
\hline
\end{tabular}
\caption{\textbf{Gender: empirical probabilities of possessing at least one latent feature.} These probabilities are directly computed from the inferred IBP matrix $\Zb$. }\label{table:probsSex}
\end{table}
 \begin{table}[h]
 \vspace*{-3mm}
\small
\centering
\begin{tabular}{|c|c|c|c|c|c|c|c|} \hline
Patterns & (000) & (010) & (001) & (100) &  (111) & (011) & (110) \\ \hline\hline
Empirical Prob. & $0.8615$ & $0.0427$ & $0.0414$ & $0.0298$  &  $0.0023$ & $0.0022$ & $0.0020$ \\ \hline
\end{tabular}
 \vspace*{-2mm}
\caption{\textbf{Gender: Empirical probability of feature pattern activations.} These probabilities are computed directly from the inferred IBP matrix $\Zb$.}\label{tab:patternsGender}
\vspace*{-3mm}
\end{table}

\begin{figure}[h]
\centering
\subfloat[Probability of suffering from each disorder]
{\includegraphics[width=1\textwidth]{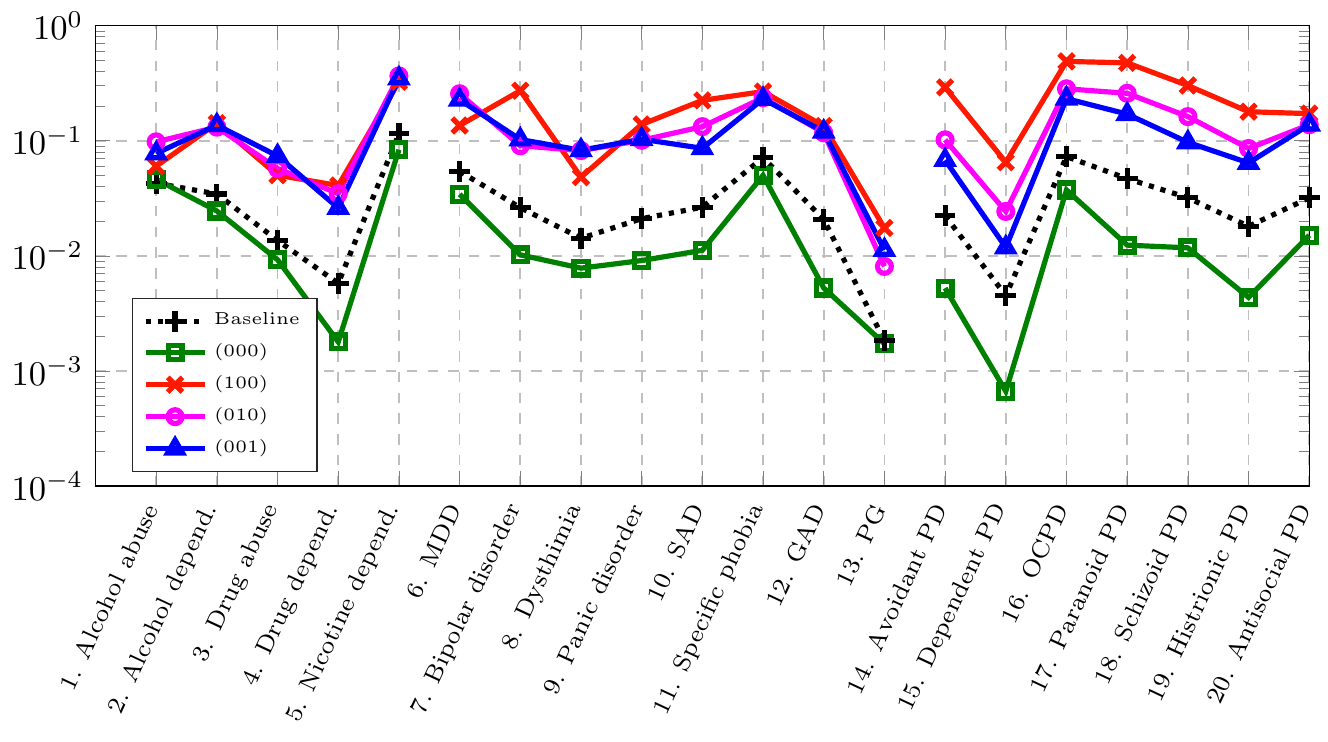}\label{fig:20Q+_4}}\\
\subfloat[Age]
{\includegraphics[width=1\textwidth]{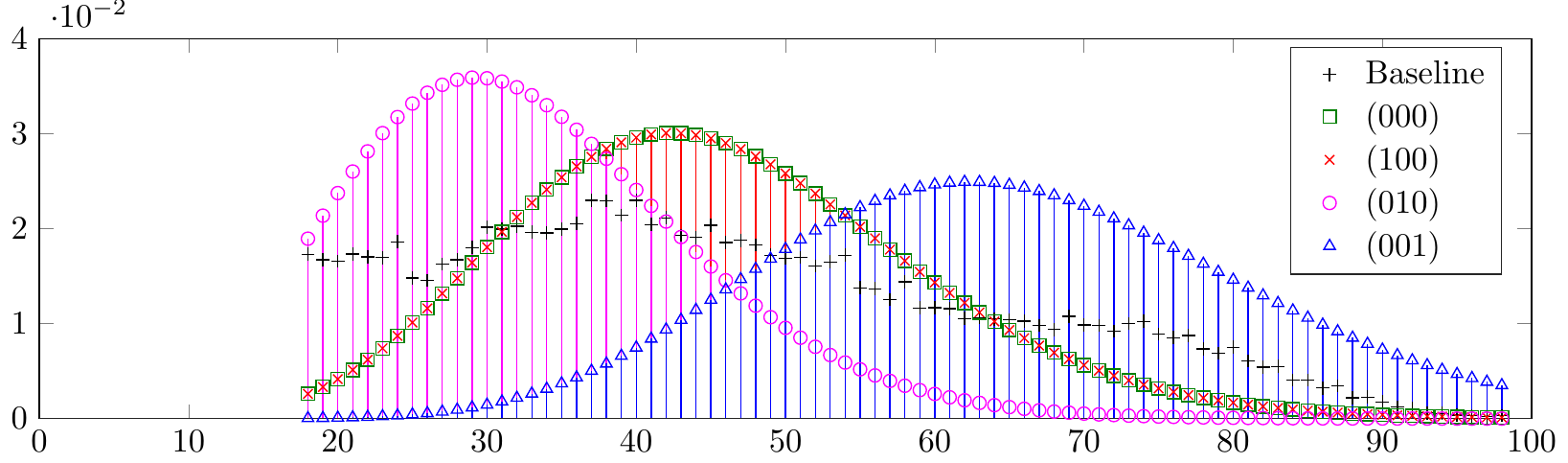} \label{fig:21Q_4}} 
\caption{\textbf{Feature effects including age in the analysis.} (a) Probabilities of suffering from the $20$ considered disorders and (b) distribution of the age for the latent feature vectors $\zn$ shown in the legend and baseline probability distribution.}
\label{fig:age}
\end{figure}

\xhdr{Age}
Now, we focus on the age of the participants, which we model as count data. 
After running our inference algorithm with the diagnoses of the $20$ disorders and the age of the subjects as input data, we again obtain three latent features, whose empirical probabilities are listed in Table~\ref{tab:probsAge}. 
Moreover, Table~\ref{tab:patternsAge} shows empirical probability of each feature pattern in the dataset. Here, we observe that the three latent features activate mostly in isolation, being the combination of two features rare (below ~3\%). 

Figure~\ref{fig:20Q+_4} shows the probability of meeting each diagnostic criteria for the latent feature vectors $\zn$ listed in the legend and in the database (baseline). In addition to the baseline probability distribution, we plot in Figure~\ref{fig:21Q_4} the inferred probability distributions over the age when none or only one of the latent variables is active (which correspond to the most common feature patterns). 
The empirical probability distribution over the age based on the data is shown (and denoted by `baseline')  in Figure~\ref{fig:21Q_4}.
Here, we observe that introducing the age of the participants as an input variable has changed the inferred latent features (with respect to the features in  \cite[]{PsiquiatrasJMLR} depicted in Figure~\ref{fig:20Q+_5}). In particular, we observe that the obtained latent features mainly differ in the probability of suffering from personality disorders (i.e., disorders from 14 to 20), being the probability of suffering from disorders 1 to 13 similar for the three plotted latent feature patterns.  In this figure, we observe that the vector $\zn$ with no active latent features, e.g., pattern $(000)$, is trying to capture the mean of the age in the database (which coincides with middle-aged subjects, i.e., $30-50$ years old). Moreover, we observe that the subjects with the highest probability of suffering from personality disorders -- pattern $(100)$ -- are likely to be middle-aged, followed in a decreasing order by young adults -- pattern $(010)$ -- and elderly people -- pattern $(001)$. 
Additionally, if we focus on the differences among the three features in disorders from 1 to 13, we also observe that, while young and elderly people tend to suffer from depression, middle-aged people tend to suffer from the bipolar disorder.
Hence, based on Figure~\ref{fig:age}, we can conclude that the bipolar disorder and the seven personality disorders tend to show up in a higher extent in the mature age, while young and elderly people tend to suffer more often from depression.

\begin{table}[h]
\small
\centering
\renewcommand{\tabcolsep}{4pt}
\begin{tabular}{|c|c|c|p{10cm}|} \hline
Feature & Empirical Prob. & Main implications \\ \hline\hline
Feature F1 & $0.0332$ & Captures severe personality disorders and middle-age subjects \\
Feature F2 & $0.0550$ & Captures mid-severe personality disorders and young subjects \\
Feature F3 & $0.0569$ & Captures mid-severe personality disorders and older subjects \\ \hline
\end{tabular}
 \vspace*{-2mm}
\caption{\textbf{Age: empirical probabilities of possessing at least one latent feature.} These probabilities are directly computed from the inferred IBP matrix $\Zb$.}\label{tab:probsAge}
 \vspace*{-2mm}
\end{table}

 \begin{table}[h]
 \vspace*{-3mm}
\small
\centering
\begin{tabular}{|c|c|c|c|c|c|c|c|} \hline
Patterns & (000) & (010) & (001) & (100) &  (011) & (110) & (101) \\ \hline\hline
Empirical Prob. & $0.8615$ & $0.0522$ & $0.0503$ & $0.0294$  &  $0.0028$ & $0.0019$ & $0.0019$ \\ \hline
\end{tabular}
 \vspace*{-2mm}
\caption{\textbf{Age: Empirical probability of feature pattern activations.} These probabilities are computed directly from the inferred IBP matrix $\Zb$.}\label{tab:patternsAge}
\vspace*{-3mm}
\end{table}

\subsubsection{Voters profile in presidential election}
Finally, we apply the proposed model to understand the correlations between demographic profiles and political vote tendencies. In particular, we focus on the United States presidential election of 1992, in which three major candidates ran for the race: the incumbent Republican president George H. W. Bush, the Democratic Arkansas governor Bill Clinton, and the independent Texas businessman Ross Perot. 
In 1992, the public's concern about the federal budget deficit and fears of professional politicians allowed the independent candidacy of billionaire Texan Ross Perot to appear on the scene dramatically~\cite[]{alvarez1995economics}, to the point of even leading against the major party candidates in the polls during the electoral race\footnote{New York Times:  \url{http://www.nytimes.com/1992/06/11/us/} \\  \url{the-1992-campaign-on-the-trail-poll-gives-perot-a-clear-lead.html}}. The race ended up with the victory of Bill Clinton by a wide margin in the Electoral College, receiving 43\% of the popular vote against Bush's 37.5\% and Perot's 18.9\%~\cite[]{lacy1999vote}. 
These results are noted for being the highest vote share of a third-party candidate since 1912, even if Perot did not obtain any electoral votes~\cite[]{lacy1999vote}.

 Our primary objective in this section is to find and analyze the different types of voters' profiles, as well as which candidate each profile tends to favor. To this aim, we used the publicly available \textit{Counties database} gathering diverse information about voting results, demographics and sociological factors per counties\footnote{Database available at: \url{http://biostat.mc.vanderbilt.edu/wiki/pub/Main/DataSets}}. 
 This dataset contains information for 3141 counties. Table~\ref{tab:counties} lists the per-county attributes that we used as input for our model.

 %
\begin{table}[h]
\centering
\small{\begin{tabular}{ |p{9.5cm}|l| } \hline
Attribute description & Type of data \\ \hline \hline
State in which the county is located  & Categorical with 51 categories\\
Population density in 1992 per squared miles & Positive real data \\
 \% of white population in 1990 & Positive real data   \\
 \% of people with age above 65 in 1990 &Positive real data  \\
\% of people above 25 years old with  bachelor's degree or higher & Positive real data  \\
 Median family income in 1989 (in dollars) & Count data \\
 \% of farm population in 1990 & Positive real data   \\
 \% of votes cast for Democratic president & Positive real data   \\
 \% of votes cast for Republican president & Positive real data   \\
 \% of votes cast for Ross Perot & Positive real data   \\
 \hline
\end{tabular}}
\caption{\textbf{List of considered attributes regarding the United States presidential election of 1992.} Attributes 1 to 7 include demographic information and sociological factors, while the last three attributes summarize the percentage voting outcome in each county.}
\label{tab:counties}
\end{table}

%
 \begin{table}[th]
 \vspace*{-3mm}
\small
\centering
\begin{tabular}{|c|c| p{9.5cm} |} \hline
Feature & Empirical Prob. & Main implications \\ \hline\hline
Feature F1 & $0.4874$  &{Favors Perot, increases the probability of  white population, and decreases average income.} \\ \hline
Feature F2 & $0.2703$  & {Favors the Democrat candidate,  increases population density, and decreases family income, percentages of white population, farming and college degrees.}\\ \hline
Feature F3 & $0.2700$ & {Favors the Republican candidate and Perot,  increases the percentage of  farming, and decreases population density.}  \\  \hline
Feature F4 & $0.0411$ &  \multirow{ 2}{*}{Capture the tails of the distributions of different attributes.} \\ 
Feature F5 & $0.0372$ &  \\ \hline
\end{tabular}
\caption{\textbf{Empirical feature activation probabilities for the Counties dataset.} We show the empirical probability of possessing at least one latent feature, as well as the main implications of the activation of each feature. These are directly computed from the inferred IBP matrix $\Zb$.}\label{table:probCounties}
\vspace*{-1mm}
\end{table}
 \begin{table}[h]
 \vspace{-1pt}
\small
\centering
\renewcommand{\tabcolsep}{4pt}
\begin{tabular}{|c|c|c|c|c|c|} \hline
Patterns & (000) & (100) & (101) & (010) & (110) \\ \hline\hline
Empirical Prob. & $0.2636$ & $0.2407$ & $0.1063$ & $0.1060$ & $0.0748$  \\
\hline
\end{tabular}
\caption{\textbf{Empirical probability of pattern activation for the top-five most popular patterns.} These probabilities are computed directly from the inferred IBP matrix $\Zb$. Features F4 and F5 are always switched off, and are thus omitted from the labels.}\label{table:probCounties2}
\end{table}

\xhdr{Experimental Setup}
We run our inference algorithm with $\alpha=5$, $\sigma_B^2=1$,   $\sigma_{\theta}^2=1$ and the  mapping transformation  from the real numbers to the positive real  numbers: $f(x) = \log(w\cdot(x - \mu) + 1)$, with $\mu = \min(\xd)$ and $w=2/\mathrm{std}(\xd)$. 
In this experiment, we activate the bias term and sample the variance of the pseudo-observations for each dimension/attribute. 
A challenging aspect of this database is that the distributions of some of its attributes are heavy-tailed, leading to a large number of  latent features as output of the GLFM, whose purpose is to capture the tails of the distributions. This is not an issue for estimation and imputation of missing data, but it renders data exploration more tedious. 
To solve this limitation, we here perform an additional data preprocessing step by applying a logarithmic transformation to heavy-tailed attributes. 
In more detail, we apply the function $g_1(x) = \log(x + 1)$ for population density, median family income, and percentage of farm population. For the percentage of white population, we used the function $g_2(x) = \log((100-x) + 1)$ since the distribution has most of its density close to 100\%.

\begin{figure}[!tpb]
	\centering
\subfloat[\% of votes cast for Ross Perot]
{\includegraphics[width=1\textwidth]{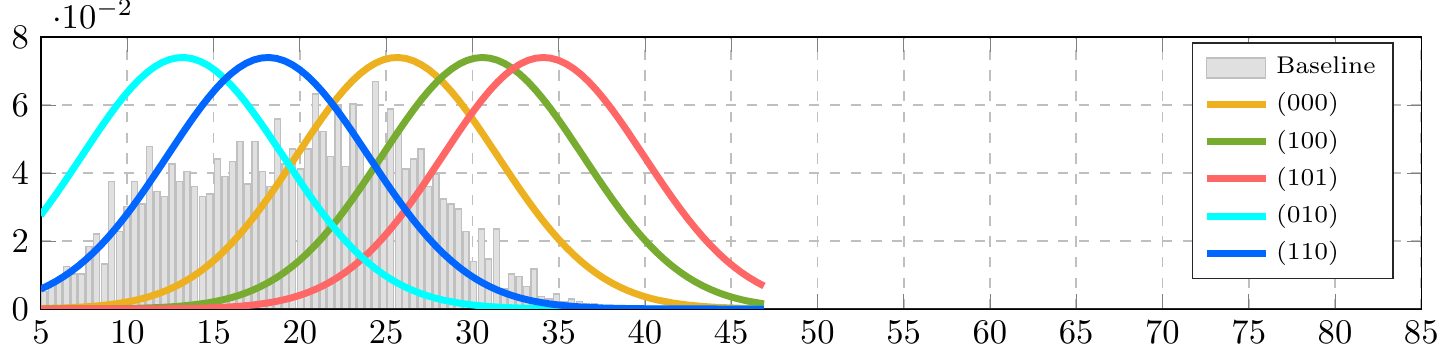}}

\subfloat[\% of votes cast for Republican candidate]{\includegraphics[width=1\textwidth]{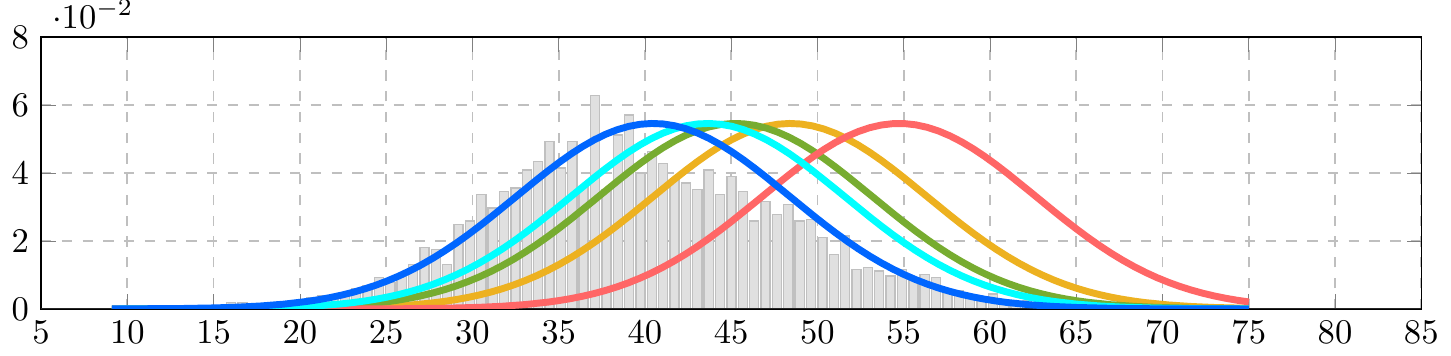}}
\vspace{1pt}
\subfloat[\% of votes cast for Democrat candidate]
{\includegraphics[width=1\textwidth]{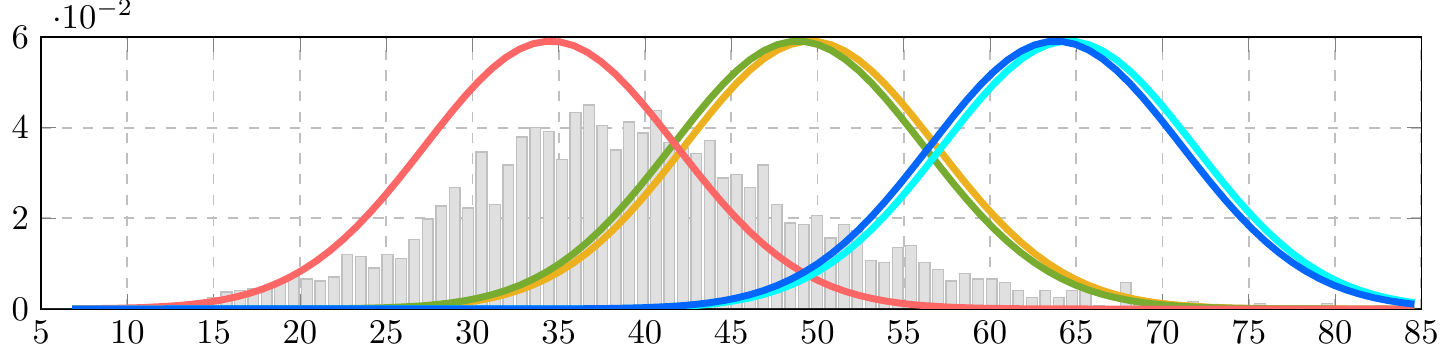}}
\vspace{1pt}
\caption{\textbf{Inferred probability distribution for the five most popular patterns.} The patterns are sorted in the legend according to their degree of popularity, as described in Table~\ref{table:probCounties2}. The baseline refers to the empirical distribution of each attribute in the entire dataset. }\label{fig:counties2}
\end{figure}

\begin{figure}[t]
	\centering
\subfloat[Pattern (000)]
{\includegraphics[width=.3\textwidth]{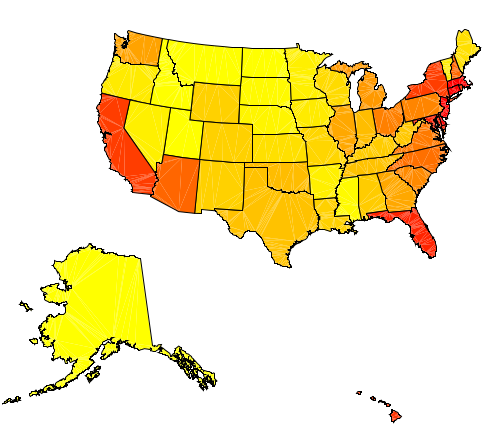}}
\subfloat[Pattern (100)]
{\includegraphics[width=.3\textwidth]{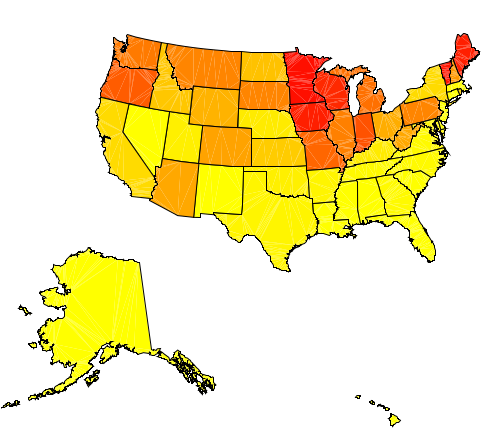}}
\subfloat[Pattern (101)]
{\includegraphics[width=.4\textwidth]{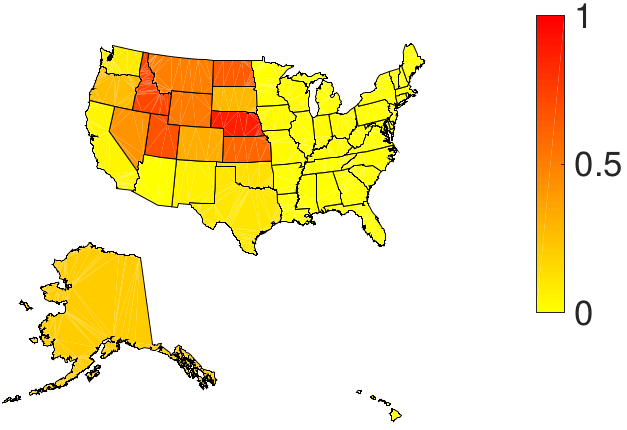}}
\vspace{3pt}
\subfloat[Pattern (010)]
{\includegraphics[width=.3\textwidth]{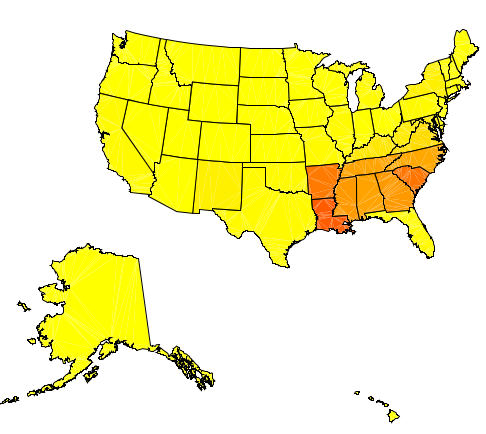}} \hspace{20pt}
\subfloat[Pattern (110)]
{\includegraphics[width=.3\textwidth]{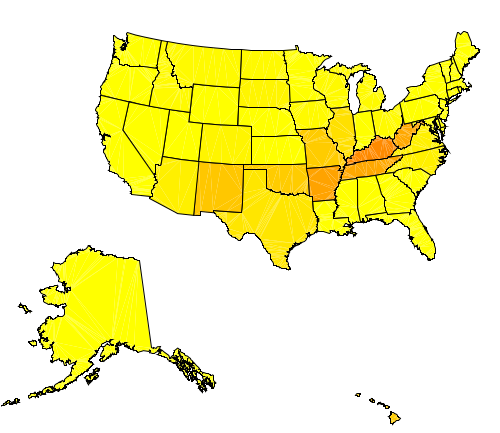}} 
	\caption{\textbf{Empirical probability of pattern activation per state.} We focus on the top-five most popular combinations of features. The label for each pattern indicates whether Features F1, F2, and F3 are active (value `1') or not (value `0'). Features F4 are F5 are always inactive in the five most common patterns, and thus are omitted in the labels.}\label{fig:counties4}
\end{figure}

\begin{figure}[t]
\vspace{1pt}
	\centering
{\includegraphics[width=.52\textwidth]{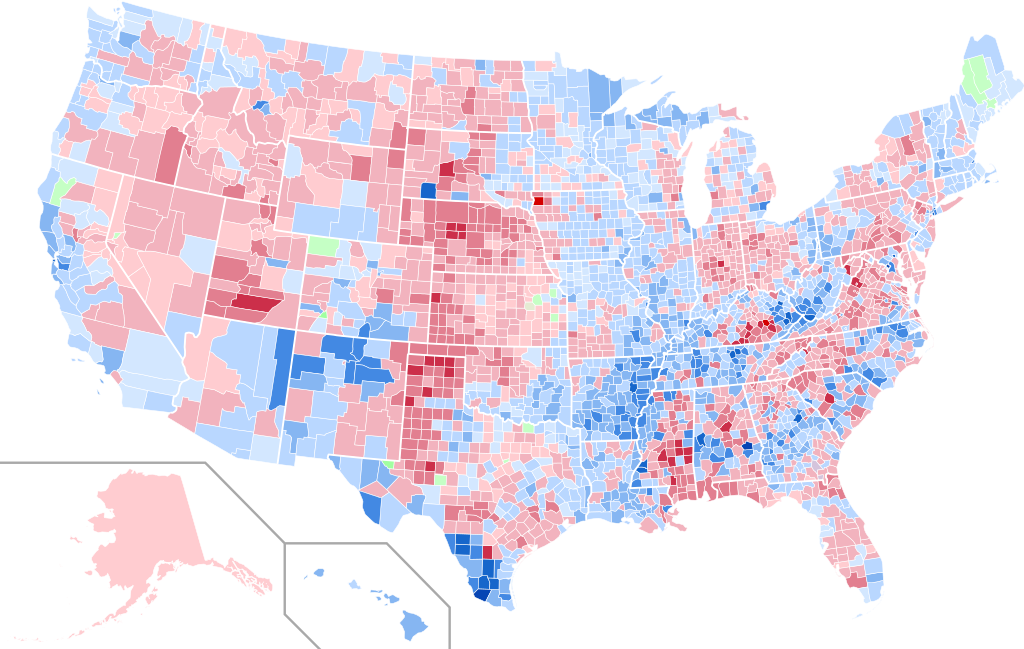}}
\vspace{3pt}
\caption{\textbf{Outcome of the 1992 presidential election per counties.} Blue color corresponds to a majority of votes for the Democrat party, red corresponds to a victory for the Republican party, green corresponds to a victory of the independent party of Ross Perot.}\label{fig:counties5}
\end{figure}

\begin{figure}[!tpb]
	\centering	
\subfloat[Population density (inhabitants/miles$^2$)]
{\hspace{-7pt}\includegraphics
{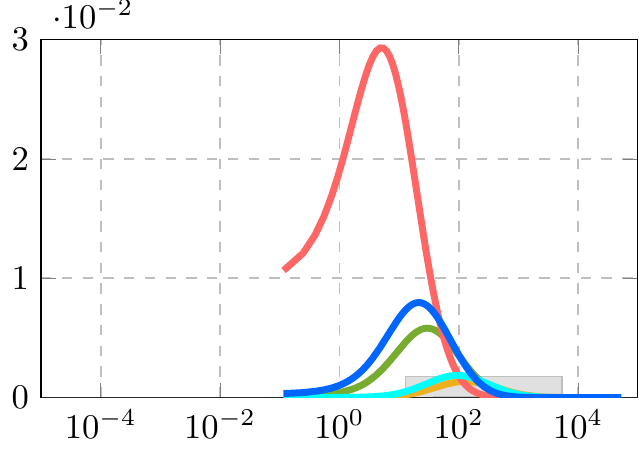}}
\subfloat[Percentage of white population]
{\includegraphics
{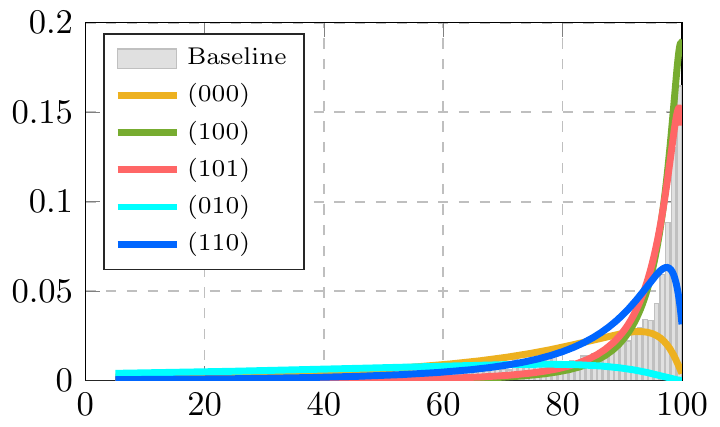}}

\begin{tabular}{p{0.5\textwidth} p{0.5\textwidth}}
\subfloat[Percentage of people $\geq$ 65 years old]
{\hspace{20pt}\includegraphics
{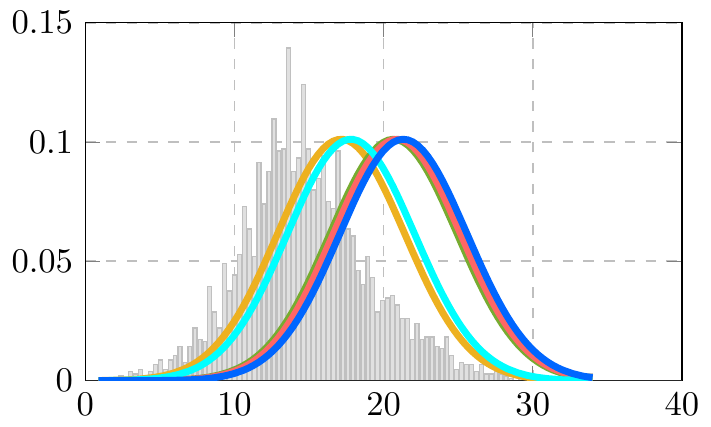}\vspace{20pt}}

&
\subfloat[Average income (in dollars)]
{\hspace{-20pt}\includegraphics
{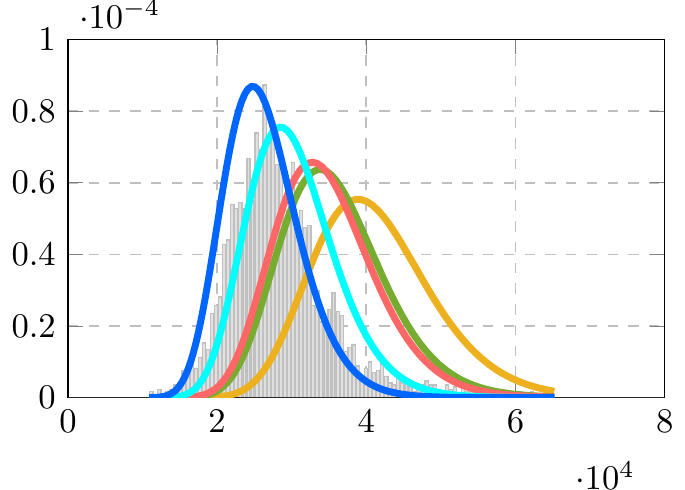}}
\end{tabular}
\subfloat[Percentage of college degrees]
{\hspace{-10pt}\includegraphics
{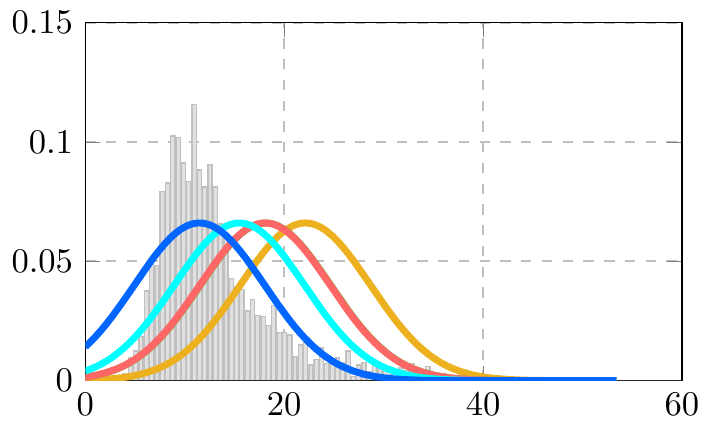}\label{fig:college}}\subfloat[Percentage of farming population]{\includegraphics
{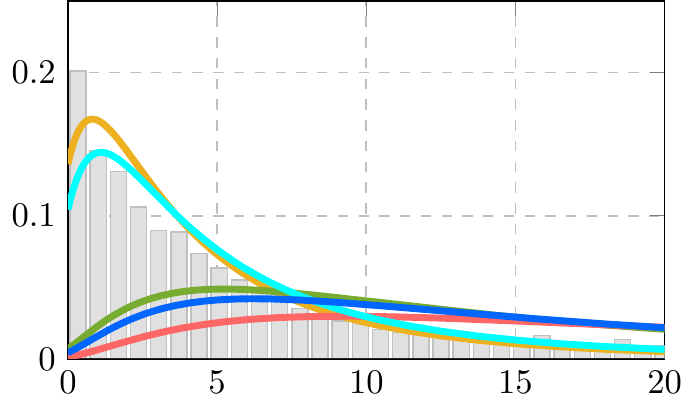}}
\vspace{1pt}
\caption{\textbf{Inferred probability distribution for the most occuring patterns.} The baseline refers to the empirical distribution of each attribute in the whole dataset.}\label{fig:counties3}
\end{figure}

\xhdr{Results}
Running the GLFM on this data results in  5 latent features, whose empirical activation probabilities are shown in Table~\ref{table:probCounties}. 
%
%
%
 Here, we observe that while the first three features are active for at least 27\% of the counties, the last two features are active only for around 4\% of the counties. 
 Moreover, we find that the different combinations of the three first latent features represent more than 92\% of the counties in USA. In the following, we will thus focus on the analysis of the three first features and, in particular, on the top-five most popular feature patterns. We show in Table~\ref{table:probCounties2}  the  empirical probabilities of these five patterns, which represent around 80\% of the U.S. counties. 
%
%
Figure~\ref{fig:counties2} shows the distribution of vote percentage per candidate associated to each of these top-five patterns, while Figure~\ref{fig:counties4} shows the corresponding geographic distribution (i.e., the empirical activation probability) across states for each of these patterns. 
In these figures, we observe that:
\begin{compactitem}
	\item[(i)] pattern (000), corresponding to the bias term, tends to model middle values for the percentage of votes for the three candidates (with an average percentage of votes of $\sim50\%$ for the Democrat candidate, $\sim48\%$ for the Republican candidate and $\sim27\%$ for Perot), and activates mainly in the east and west coasts of the country, as well as Florida;
	\item[(ii)] pattern (100) provides similar percentage of votes for the Democrat and Republican candidates  as in pattern (000), but it favors the independent candidate Perot (with an average percentage of votes above 30\%), and activates mostly in the north central-east region of the country and Maine (the state where Perot's party managed to beat the Republican party);
	\item[(iii)] pattern (101) activates in the north central-west region of the USA (not including the coast) and represents a profile inclined towards the Republican party (with an average percentage of votes of  $\sim55\%$) while also favoring in a lower extent the independent candidate; and
	\item[(iv)] patterns (010)  and (001) clearly capture Democrat-oriented profiles, and activate mainly in the south east region of the USA, including the state from which Bill Clinton comes from, Arkansas. 
\end{compactitem}
Note that the demographic results above are in agreement with the outcome of the election per counties\footnote{\url{https://en.wikipedia.org/wiki/United_States_presidential_election,_1992}}, as shown in Figure~\ref{fig:counties5}. 
Next, we analyze the demographic information associated to each of the feature patterns above. To this end, we show  in  Figure~\ref{fig:counties3} the distribution of each attribute/dimension of the data for each of the considered patterns. 
%
%
First, we observe that pattern (000), which activates mostly  in the coasts and Florida, corresponds to the highest population density, average income, and percentage of college degrees, as well as an important race diversity and low farming activity. These observations align with the typical profile characterizing ``big-cities''. As stated before, this pattern is the most balanced in terms of voting tendency, with an equilibrated support for both Democrat and Republican, as well as intermediate values for the percentage of votes cast for Perot.

Second, patterns (100) and (101) represent the largest share of Perot's votes, both with an average percentage of votes above 30\% for Perot. 
Figure~\ref{fig:counties3} shows that Perot's main supporters, characterized mainly by pattern (101), also correspond to Republican main supporters, who tend to live in low populated areas in the north central part of the country where farming activity is considerable, and the percentages of white population and over-65 years old population are also high. 
The second voting force backing Perot, captured by pattern (100) and located in the north east-central part of USA, corresponds mostly to white population with an intermediate-high average income and an average percentage of college degrees around 18\% (the red curve in Figure~\ref{fig:college} overlaps the green line). 
These results back the analysis in \cite[]{lewis1994}, which showed that the majority of Perot's voters (57\%) were middle class, earning between \$15,000 and \$49,000 annually, with the bulk of the remainder drawing from the upper middle class (29\% earning more than \$50,000 annually).
 Perot's campaign ended up taking 18.9\% of the votes, finishing second in Maine and Utah, as captured by pattern (100) and (101) respectively. 

 Finally, Democrat's patterns (010) and (110) are  mainly active in the Southeastern United States, and capture a diverse range of voters in terms of their demographic properties. On the one hand,  pattern (010)  captures highly populated counties, with low values of family income, percentage of college degrees, percentage of white population  and percentage of farming population. 
 %
 %
On the other hand, pattern (110) captures  low populated counties with a  large percentage of population above 65 year old, as well as a larger presence of farming activity and lower average income. 
 %
 %
These results might be explained by the broad appeal across all socio-ethno-economic demographics that the Democratic party has historically targeted. %

\section{Conclusions}
\label{sec:conclusions}

In this paper, we developed an efficient general latent feature model, named GLFM,  suitable for modelling and inference with real-world heterogeneous datasets. The proposed model presents attractive properties in terms of flexibility and interpretability. 
Its nonparametric nature allows it to automatically infer the appropriate model complexity (i.e., number of latent features) from data. 
Secondly, the fact that the latent features are binary-valued makes it easier to identify and interpret meaningful patterns in the data exploration process. 
Thirdly, the model shares the properties of conjugate models, which allow us to  propose an efficient inference scheme that scales linearly with the number of observations (objects) and dimensions (attributes) in the data.

We showed the flexibility and applicability of the proposed GLFM by solving both prediction and data exploration tasks in several real-world datasets. 
In particular, we used the proposed model to estimate and replace missing data in heterogeneous databases.
Also, we used the GLFM for data exploratory analysis of real-world datasets related to diverse application domains: clinical trials, psychiatry, sociology and politics. The software library provides the first available software for latent feature modeling in heterogeneous data, and includes user-friendly functions for missing data estimation and data exploratory analysis. 
The software package is publicly available at: \url{https://github.com/ivaleraM/GLFM} and reviewed in Appendix~\ref{App:Toolbox}.
We hope to bring a powerful tool for researchers from diverse knowledge domains  to help them analyze a wide range of datasets in an automatic manner.

\section{Acknowledgements}
Isabel Valera acknowledges her Humboldt Research Fellowship for Postdoctoral Researchers. Maria Lomeli  and Zoubin Ghahramani acknowledge support from the Alan Turing Institute (EPSRC Grant EP/N510129/1) and EPSRC Grant EP/N014162/1, and donations from Google and Microsoft Research. 
Melanie F. Pradier is grateful to the European Union 7th Framework Programme through the Marie Curie Initial Training Network ``Machine Learning for Personalized Medicine'' MLPM2012, Grant No. 316861 (http://mlpm.eu).
This work has been partly supported by MINECO/FEDER ('ADVENTURE',
 id. TEC2015-69868-C2-1-R), and Comunidad de Madrid (project
 'CASI-CAM-CM', id. S2013/ICE-2845). 

{

\bibliographystyle{plain}
\bibpunct{[}{]}{,}{n}{}{;}
\bibliography{Bib}
}
\appendix
\section*{GLFM: General Latent Feature Modeling Toolbox}\label{App:Toolbox}
\section{Introduction}
\vspace*{-5pt}
The GLFM is a general Bayesian nonparametric latent feature model suitable for data exploration or missing values imputation in heterogeneous datasets, where the attributes describing each object can be either discrete, continuous or mixed variables. 
Specifically, the GLFM accounts for real-valued and positive real-valued data as examples of continuous variables, and categorical, ordinal and count data as examples of discrete variables. 

The GLFM builds on the Indian Buffet Process~\cite[]{IBP}, and therefore, it assumes that each observation $x_n^d$ can be explained by  a potentially infinite-length binary vector $\mathbf{z}_n$ whose elements indicate whether a latent feature is active or not for the $n$-th object; and a (real-valued) weighting vector $\mathbf{B}^d$, whose elements weight the influence of each latent feature in the $d$-th attribute\footnote{For agreement with the notation in~\cite[]{ourPaper}, we here capitalize the vector $\mathbf{B}^d$.}. 
%
Since the product of the latent feature vector and the weighting vector leads to a real-valued variable, it is necessary to map this variable to the desirable output (continuous or discrete) space, for example, the positive real line or the finite ordered set $\{$\textit{low, medium, high}$\}$. Thus, the GLFM assumes the existence of intermediate real-valued auxiliary variables $y_n^d\sim \mathcal{N}(\mathbf{z}^n\mathbf{B}^d, \sigma^2_d)$, called pseudo-observations, and a transformation function $f_d(\cdot)$ that maps this variable into the actual observation $x_n^d$, i.e., $x_n^d = f_d(y_n^d+u)$ where $u\sim \mathcal{N}(0, \sigma^2_u)$ is an auxiliary noise with zero mean and small variance $\sigma^2_u$. 
Figure~\ref{fig:IBPdiscreto} illustrates the GLFM for a ordinal attribute taking values in the ordered set $\{$\textit{low, medium, high}$\}$. 

\begin{figure}[t]
\vspace*{-35pt}
\includegraphics[width=0.7\textwidth]{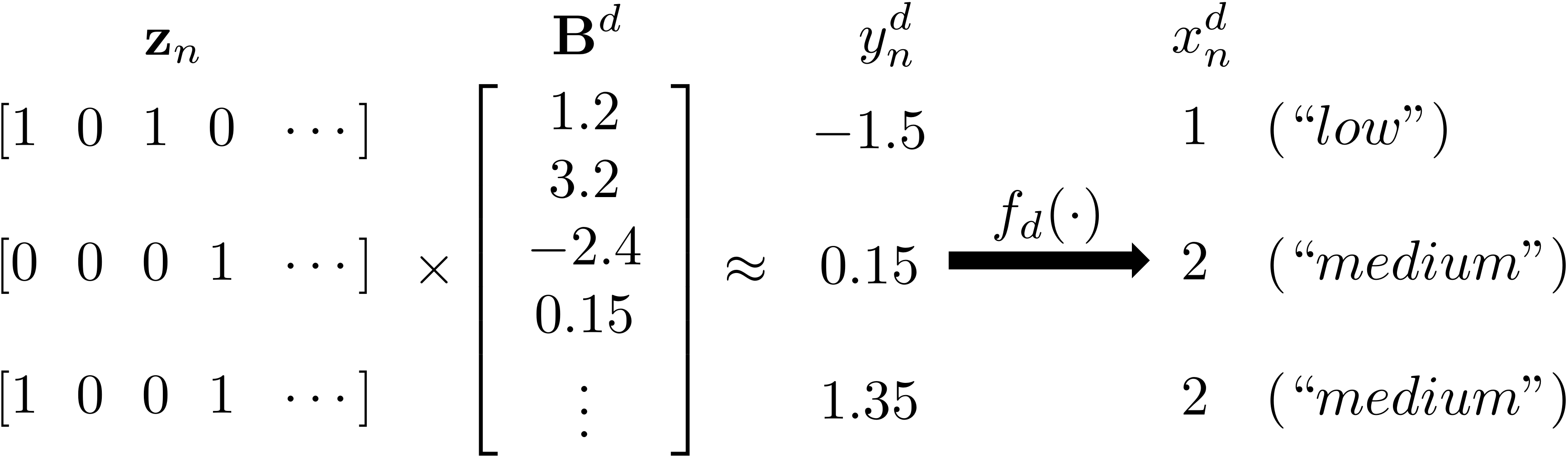}
\centering
\vspace{-1pt}
\caption{{Illustration of the GLFM.}}
\label{fig:IBPdiscreto}
\vspace{-.6cm}
\end{figure}

\vspace*{-10pt}
\section{Implementation}
\vspace*{-5pt}
The GLFM package contains an efficient C++ implementation, together with an user interface in Python, Matlab and R, of the collapsed Gibbs sampling algorithm described in  Algorithm~1. The main function of the package, \texttt{hidden = GLFM.infer(data)},\footnote{This call corresponds to a python call. The equivalent call in Matlab is hidden = GLFM\_infer(data) and in R output $\leftarrow$ GLFM\_infer(data).} runs the inference algorithm given the input structure \texttt{data} and returns the learned latent variables in the output structure \texttt{hidden}.
This function receives as input an observation matrix $\mathbf{X}$ and a vector indicating the type of data for each dimension (optionally, model hyper-parameters and simulation settings can be customized by the user). The latent variables are learned by making use of the mapping transformations listed in Table~\ref{table:functions} to account for the continuous and discrete data types mentioned above. 
Here, the parameters $\mu$  and $w$ are used to shift and scale the raw input data, and are respectively set to the empirical mean and the standard deviation  for real-valued attributes, and to the minimum value and the standard deviation for positive real-valued and count attributes. 
This guarantees that the prior distributions on the latent variables are equally good for all the attributes in the dataset, regardless their support.
%
The output structure \texttt{hidden} contains the latest MCMC sample of the  latent feature vectors $\mathbf{z}^n$ for $n=1,\ldots,N$, the weighting vectors $\mathbf{B}^d$, as well as auxiliary variables (including the pseudo-observation variances $\sigma^2_d$ and the thresholds $\theta_r$) necessary for the corresponding transformation $f_d(\cdot)$, for $d=1,\ldots,D$.

%
%
%

Moreover, since the Bayesian nonparametric nature of the GLFM allows the model complexity (i.e., the length of the vectors $\mathbf{z}^n$ and $\mathbf{B}^d$) to grow with the number of observations, we bound the model complexity to a finite value, which is an additional input to the GLFM package. 
This bound allows us not only to keep the model complexity, and therefore the running time, under control, but also to efficiently manage the memory allocation. 
Finally, our implementation of the GLFM makes use of the GNU Scientific Library (GSL),\footnote{\url{https://www.gnu.org/software/gsl/}} to efficiently perform a large variety of mathematical routines such as random number generation, and matrix or vector operations. 

\begin{table}[t]
\vspace*{-30pt}
\centering
\small
\begin{tabular}{cll}
Type of Variable & Domain & Transformation $x = f_d(y)$  \\\hline \hline  \vspace{10pt}

Real-valued & $x \in \Re$ & $ x= w (y+u)+\mu$  \\  \vspace{10pt}

Positive real-valued & $x \in \Re^+$ & $x=\log(\exp(w(y+u) + \mu)+1)$  \\ \hline  \vspace{10pt}

Categorical & $x \in \{1, 2, \ldots, R\}$ (unordered set) & $x= \arg\max_{r\in\{1,\ldots, R\}} y_{r}$ \\  \vspace{10pt}
Ordinal & $x \in \{1, 2, \ldots, R\}$ (ordered set)& $x=\left\{
  \begin{array}{l l}
    1 & \quad \mathrm{if }\quad y \leq\theta_1\\ 
    2 & \quad \mathrm{if } \quad \theta_1 <y \leq\theta_2\\
     & \vdots \\
    R & \quad \mathrm{if } \quad \theta_{R-1} <y\\ 
  \end{array} \right.$ 
  \\ 
Count & $x \in \{1, 2, 3,\ldots\}$ & $x=\lfloor \log(\exp(w(y+u) + \mu)+1) \rfloor$  \\ \hline
\end{tabular}
\vspace{-1pt}
\caption{Mapping functions implemented in the toolbox. }\label{table:functions}
\vspace{-.5cm}
\end{table}

\vspace*{-10pt}
\section{Usage}
\vspace*{-5pt}

\subsection{Data preprocessing and initialization}
\vspace*{-5pt}
A convenient property of the GLFM package is that it can be used blindly on raw data without requiring any preprocessing step on the dataset, nor special tuning of hyperparameters.
The only requirement for the user to use the package is to format the data as a numerical matrix of size $N\times D$ and indicate in an additional vector the type of data for each of the $D$ attributes.  
As mentioned above, the parameters of the transformations in Table~\ref{table:functions} are internally fixed to ensure that the pseudo-observation for all attributes fall in a  similar interval of the real line, so that the prior distribution on the latent variables is equally good for all attributes.

\begin{figure}[t]
\centering
\vspace*{-30pt}
\subfloat[$\mathbf{X}^d$ without preprocessing]
{\includegraphics[width=0.33\textwidth]{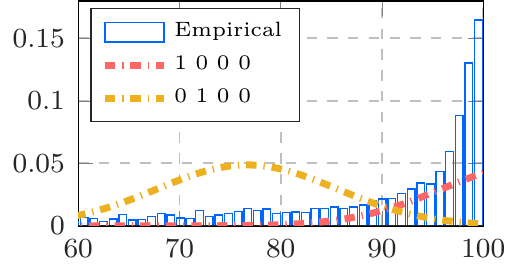}\label{fig:T2}}
\subfloat[$\mathbf{X}^d$ with log preprocessing]
{\includegraphics[width=0.33\textwidth]{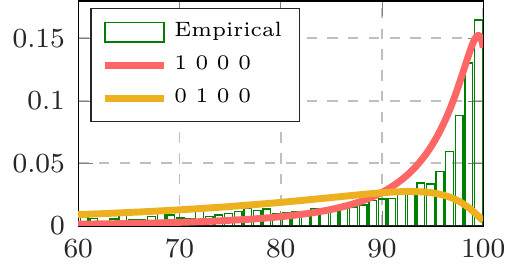}\label{fig:T3}}
\subfloat[Pseudo-observations $\mathbf{y}^d$]
{\includegraphics[width=0.295\textwidth]{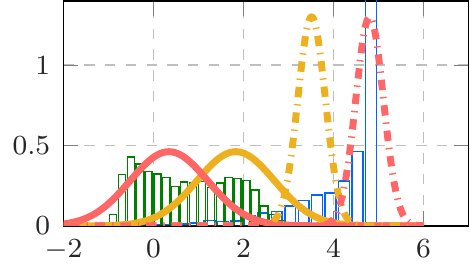}\label{fig:T1}}\hspace{0.1cm}
\vspace{-1pt}
\caption{Illustration of optional data preprocessing. Panel (a) and (b) show the histograms of, respectively, a heavy-tailed attribute  and the attribute after a logarithmic transformation, as well the distribution of the inferred latent feature patterns. Panel (c) shows the histogram of the pseudo-observations inferred for the original and the preprocessed attributes, as well the distribution of the inferred latent feature patterns. Here, we observe that the distribution of the attribute is better captured by the latent model when a preprocessing step is performed to correct/minimize the non-Gaussian behavior  of the attribute.  
}
\label{fig:transform}
\vspace*{-.5cm}
\end{figure}

However, we incorporate an additional functionality that allows the user to specify external preprocessing functions to further improve the performance of the algorithm. For instance, in cases in which the distribution of an attribute presents a clearly non-Gaussian behavior, e.g., it is concentrated around a single value or heavy-tailed, it might be suitable to preprocess this variable by applying a logarithmic transformation, as shown Figure~\ref{fig:transform}.
%
%


\vspace*{-10pt}
\subsection{Missing Data Estimation}
\vspace*{-5pt}
GLFM can be used for estimation and imputation of missing data in heterogeneous datasets, where the missing values can be encoded with any (numerical) value that the user specifies. The Bayesian nature of the GLFM allows to efficiently infer the latent feature representation of the data using the available information (i.e., the non-missing values), and using it to compute the posterior distribution of each missing value in the data. 
Note that given the posterior distribution of each missing value, one might opt for different approaches to impute missing values, e.g., one might opt for imputing a sample of the posterior distribution or simply the maximum a posteriori (MAP) value.  
The GLFM package provides the function \texttt{[Xmap, hidden]=GLFM.complete(data)} which  infers the latent feature representation, given the (incomplete) observation matrix, and returns a complete matrix where the missing values have been imputed to their MAP value; and  the hidden structure containing all the inferred latent variables. 
This function therefore runs the inference function \texttt{GLFM.infer()}, as well as the function \texttt{GLFM.computeMAP()}, which computes the MAP of a single missing element $x_n^d$ given $\mathbf{z}^n$ and $\mathbf{B}^d$. 

\vspace*{-10pt}
\subsection{Data Exploration Analysis}
\vspace*{-5pt}
GLFM can also be used as a tool for  data exploratory analysis, since it is able to find the latent structure in the data and capture the statistical dependencies among the objects and their attributes in the data. 
GLFM provides (weighted) binary latent features, easing their interpretation and making it possible to cluster the objects according to their activation patterns of latent features. 
Moreover, it also allows to activate a latent feature that is active for all the objects  (a.k.a. bias term), which might be useful to capture the mode of the distribution of each attribute in the dataset. 
In order to ease data exploration, GLFM provides the function \texttt{GLFM.plotPatterns()}, which 
plots the posterior distribution for of each attribute under the given latent feature  patterns, and therefore, allows us to find patterns and dependencies across both objects and attributes.  This function, in turn, makes use of the function  \texttt{GLFM.computePDF()}, which evaluates the posterior distribution of an attribute under a given latent feature vector.

\vspace*{-10pt}
\subsection{Examples}
\vspace*{-5pt}
The package manual contains simple examples demonstrating package usage. Additionally, we provide the following demonstrations (with the scripts in Python, Matlab and R):
\begin{compactitem}
\item demo toy example: Simple illustration of GLFM pipeline, replicating the example of the IBP linear-Gaussian model in~\cite[]{IBP}.
\item demo completion: Illustration of missing data estimation on the MNIST image dataset.
\item demo data exploration (counties \& prostate): Replication of results on data exploration in the main paper. This demo requires data download, which is instructed.
\end{compactitem}

\vspace*{-10pt}
\section{Availability and Documentation}
\vspace*{-5pt}
GLFM code is publicly available in \url{https://github.com/ivaleraM/GLFM}, where we provide  a technical
document introducing the model and a user manual describing the usage details of the toolkit, including software requirements. The Python and Matlab implementations are under MIT license. The R implementation extends the \textit{RcppGSLExample}\footnote{\url{https://github.com/eddelbuettel/rcppgsl/tree/master/inst/examples/RcppGSLExample}}, and therefore, is under GPL ($>=$ 2) license.


%
%

\end{document}